\pdfoutput=1

\documentclass[11pt]{article}

\usepackage{ACL2023}

\usepackage{times}
\usepackage{latexsym}
\usepackage{graphicx}
\usepackage{booktabs}
\usepackage{adjustbox}
\usepackage{multirow}
\usepackage{amsmath}
\usepackage{amssymb}
\usepackage{listings}
\usepackage{xcolor}
\usepackage{hyperref}
\newcommand{\method}{\textsc{Method}}

\usepackage{makecell}

\usepackage{subcaption}
\usepackage[T1]{fontenc}
\usepackage[utf8]{inputenc}
\usepackage{microtype}
\usepackage{inconsolata}

\usepackage{amsmath}
\usepackage{graphicx}
\usepackage{url}
\usepackage{xcolor}
\usepackage{enumitem}

\newcommand{\diagonality}{\textsc{C-Dist }}

\title{From Surveys to Narratives: Rethinking Cultural Value Adaptation in LLMs}

\author{
M. Farid Adilazuarda$^{1}$, Chen Cecilia Liu$^{2}$, Iryna Gurevych$^{2}$, Alham Fikri Aji$^{1}$
\\
        \textsuperscript{1}MBZUAI \\ 
        \textsuperscript{2}Ubiquitous Knowledge Processing Lab (UKP Lab), \\
        Technical University of Darmstadt and \\
        National Research Center for Applied Cybersecurity ATHENE, Germany \\ 
}

\begin{document}
\maketitle

\begin{abstract} 
Adapting cultural values in Large Language Models (LLMs) presents significant challenges, particularly due to biases and limited training data. Prior work primarily aligns LLMs with different cultural values using World Values Survey (WVS) data. However, it remains unclear whether this approach effectively captures cultural nuances or produces distinct cultural representations for various downstream tasks. In this paper, we systematically investigate WVS-based training for cultural value adaptation and find that relying solely on survey data can homogenize cultural norms and interfere with factual knowledge. 
To investigate these issues, we augment WVS with encyclopedic and scenario-based cultural narratives from Wikipedia and NormAd. While these narratives may have variable effects on downstream tasks, they consistently improve cultural distinctiveness than survey data alone. Our work highlights the inherent complexity of aligning cultural values with the goal of guiding task-specific behavior. We release our code at \url{https://github.com/faridlazuarda/from-surveys-to-narratives}.

\end{abstract}

\section{Introduction}

Recent research in Large Language Models (LLMs) suggest LLMs align closely with the cultural values of Western, Educated, Industrialized, Rich, and Democratic (WEIRD, \citealt{henrich2010weirdest}) societies without adaptations \cite[among others]{DBLP:journals/corr/abs-2203-07785, nvm/acl/ramezani-xu-2023-knowledge, analysis/cao-etal-2023-assessing}. The WEIRD-centric bias can harm specific groups and limit the model's usefulness to a diverse global audience. Indeed, culture is a distinct and vital aspect of human society, influencing behavior, norms, and worldviews \citep{geertzinterpretation}. However, current research lacks robust mechanisms to adapt LLMs outputs in ways that reflect different cultural value systems (i.e., culturally adapt LLMs).\footnote{For this paper, we focus on ``culture'' at a linguistic-regional level (e.g., Iraq and Jordan represent \textbf{Arab} culture vs.\ Argentina and Mexico that represent \textbf{Spanish} culture), but we acknowledge that culture is more nuanced, including sub-cultures within a group and intersectional factors such as ethnicity and religion \cite{adilazuarda-etal-2024-towards}.}

\begin{figure}[t]
    \centering
    \includegraphics[width=1\linewidth]{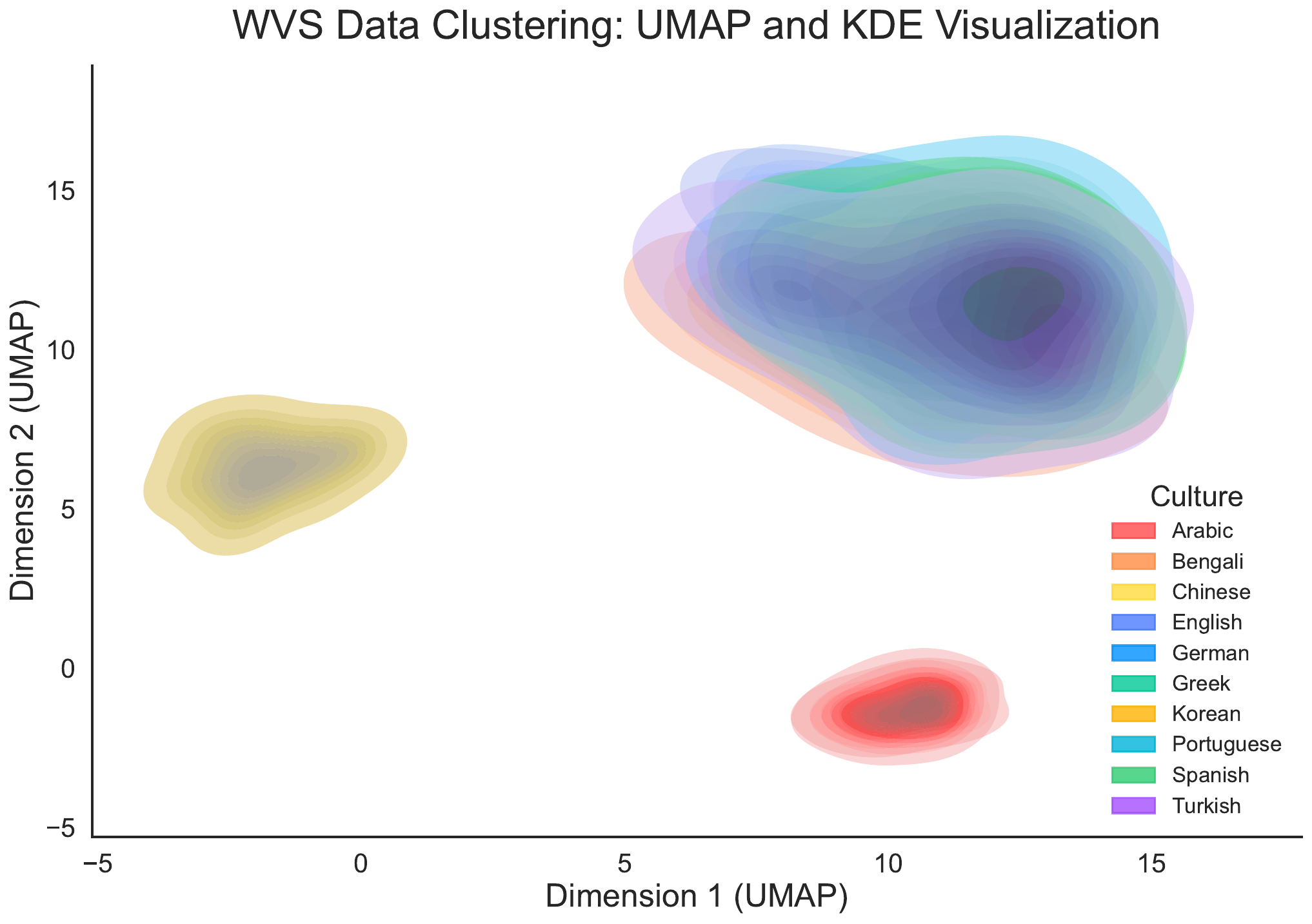}
    \caption{UMAP-KDE visualization of cultural value distributions from WVS data reveals significant homogenization. While Arabic (lower right) and Chinese (left) cultures form distinct clusters, many others converge in the upper right. This suggests that current WVS-based training may be insufficient to capture cultural nuances.}
    \label{fig:wvs_clustering}
\end{figure}

Existing work often adapts LLMs to cultural values by leveraging self-reported survey data~\cite{cultureLLM/corr/abs-2402-10946, DBLP:journals/corr/abs-2410-12971, culturePark/corr/abs-2405-15145} such as the World Values Survey (WVS, \citealt{wvs}). Although WVS offers a quantitative glimpse into cultural attitudes (e.g., \textit{“How important is family in your life?”} on a scale from 1 to 4), it remains unclear how to best translate these numeric indications into concrete behavior in downstream tasks (e.g., classification of offensiveness in different linguistic-cultural settings). Beyond survey responses on values and opinions, culture also includes social norms, historical contexts, and nuanced beliefs \cite{DBLP:journals/corr/abs-2406-03930} that may not be fully captured through self-reported questionnaires. As shown in Figure~\ref{fig:wvs_clustering}, even WVS data for distinct cultures may converge into overlapping clusters in latent space (showing semantic similarities), potentially homogenizing nuanced cultural dimensions.

Ideally, cultural value adaptation should also enhance downstream tasks within each culture. However, several challenges emerge. First, adapting multiple cultural values may create interference similar to that seen in multilingual models \cite{cursexling/conneau-etal-2020-unsupervised, cursexling/wang-etal-2020-negative}, given language-culture interconnections \cite{adilazuarda-etal-2024-towards, hershcovich-etal-2022-challenges, hovy-yang-2021-importance}. 
Second, the reliability of cultural value training data is uncertain. Studies show discrepancies between attitude and actual behavior in human \cite{gross1975attitude, fazio1981direct}, raising concerns about the WVS's ability to accurately reflect cultural behavior for LLM training, necessitating further investigation. 

In this work, we tackle these challenges through a critical evaluation of current cultural value adaptation methods. Through a series of experiments, we reveal the key limitations of using WVS as training data: while WVS provides insights into cultural values, it lacks the contextual depth needed to inform value-driven behavior in downstream tasks. Given these limitations of survey data, we investigate whether augmenting WVS with richer narrative sources like encyclopedic descriptions (Wikipedia) and scenario-based norms (NormAd; \citealt{rao2024normadframeworkmeasuringcultural}) yields more culturally distinct and effective LLM adaptations. We assess the impact on preserving cultural distinctiveness, downstream task performance, and factual knowledge.

To summarize, our contributions are: \textbf{1)} We identify \textbf{\emph{cultural interference}} in adaptation using WVS, showing that it improves tasks like offensiveness classification but tends to homogenize cultural behaviors rather than preserve their differences. \textbf{2)} We demonstrate \textbf{\emph{knowledge interference}} from adaptation, where adaptation can degrade factual knowledge understanding. \textbf{3)} We investigate the impact of \textbf{\emph{augmenting WVS with context-rich narratives (Wikipedia, NormAd)}} and find that, while their effects on downstream tasks can vary, they help preserve cultural distinctiveness. 
Our analysis showcased the challenge of aligning cultural values to influence task-specific behavior and highlighted the need for further research into \emph{which types of data} best support effective cultural adaptation.

\section{Methodology}
\label{sec:methodology}
We systematically designed experiments to investigate our research question. This section details our methodologies for cultural adaptation and performance evaluation.  We begin with zero-shot prompting, followed by single-culture adapter fine-tuning, and conclude with an analysis of cross-cultural interference using auxiliary tasks such as MMLU (Massive Multitask Language Understanding; \citealt{hendrycks2021measuring}). We describe datasets, models, and evaluation metrics in \S\ref{sec:experimental_setup}.

\subsection{Zero-Shot Prompting} \label{subsec:zeroshot_prompting} \textbf{Zero-shot prompting} leverages a pre-trained LLM without additional fine-tuning. To adapt the model for a specific target culture, we use simple instructions that reference the culture. For instance, for an \textsc{OffensEval}-style task, we use the following prompt in Table~\ref{tab:prompt}:

\begin{table}[h]
    \centering
    \small
    \resizebox{\columnwidth}{!}{%
    \begin{tabular}{p{0.95\columnwidth}}
        \toprule
        You are a \texttt{\{country\}} chatbot that understands \texttt{\{country\}}'s cultural context. \\[1ex]
        \textbf{Question:} Is the following sentence offensive according to \texttt{\{country\}}'s cultural norms? \\[1ex]
        \textbf{Input:} \texttt{\{input\_txt\}} \\[1ex]
        \textbf{Answer:} [Select one: 1. Offensive, 2. Not offensive] \\
        \bottomrule
    \end{tabular}%
    }
    \caption{Zero-shot prompt template for offensiveness classification. We list the full prompts used in our study in Appendix \ref{app:prompts}.}
    \label{tab:prompt}
\end{table}

Here, the model's responses rely entirely on cultural or multilingual knowledge that was encoded during pre-training. This can create systematic biases when the training data is skewed toward dominant cultural paradigms, which may disadvantage underrepresented groups \cite{guo2024biaslargelanguagemodels}.

\subsection{Cultural Value Adaptation via Fine-tuning}
\label{subsec:finetuning_methods}

Beyond zero-shot prompts, we explore explicit fine-tuning with culture-specific data, referred to as \emph{single-culture adaptation} in our paper. Following \citet{cultureLLM/corr/abs-2402-10946}, we train a separate LoRA adapter \cite{hu2021loralowrankadaptationlarge} for each cultural context using data from a single or a combination of data sources. Each adapter is specialized to reflect the norms, attitudes, or knowledge of specific culture. However, data sparsity and overfitting are risks, particularly for cultures with limited samples.

In single-culture adaptation, each LoRA adapter is trained to reflect the high-level cultural values present in the training dataset. During inference, the appropriate adapter is activated based on the test target culture specified.

\section{Experimental Setup}
\label{sec:experimental_setup}

We base our experiments on the CultureLLM \cite{cultureLLM/corr/abs-2402-10946} framework, one of the earliest popular adaptation frameworks for cultural values. We design our experimental setup to evaluate across multiple LLMs and languages. Below, we briefly describe datasets used for training and evaluation, model and training hyperparameters, and evaluation metrics.

\subsection{Linguistic-Cultural Settings}
We conduct experiments on ten distinct linguistic-cultural settings. Here, we use the ISO 693-3 code for simplicity: 
Arabic (ara, Iraq and Jordan), Bengali (ben, Bangladesh), Chinese (zho, China), English (eng, United States), German (deu, Germany), Greek (ell, Greece), Korean (kor, South Korea), Portuguese (por, Brazil), Spanish (spa, Argentina and Mexico), and Turkish (tur, Turkey). 

\paragraph{Terminology.} In this paper, we use \textit{culture} to denote a linguistic–regional grouping (e.g., “Arab” represented by Iraq and Jordan; “Spanish” by Argentina and Mexico). \textit{Country} refers to specific national datasets used to instantiate a culture. \textit{Language} denotes the linguistic form of inputs/outputs (e.g., ara, spa). While related, these are not interchangeable; our experiments condition adapters on \emph{culture}, evaluate on country-specific test sets, and control for language.

\subsection{Training Dataset}
We established training scenarios with data drawn from three different sources:\\
\noindent\textbf{WVS.} In this setting, we use the WVS and semantically augmented data based on \citet{cultureLLM/corr/abs-2402-10946}. WVS is a survey data commonly used in social sciences, as well as a proxy for cultural values in NLP \cite{adilazuarda-etal-2024-towards}. The dataset consists of question-and-answer pairs that provide quantitative indicators of societal beliefs and attitudes (e.g., questions on family importance or religion). 

\noindent\textbf{Wikipedia.} We select Wikipedia articles with detailed knowledge, region-specific norms, social practices, and historical contexts of our defined cultures. These articles can enrich the numeric survey data with qualitative background.\footnote{See Table \ref{tab:sources_urls} for the Wikipedia pages used.}  

\noindent\textbf{NormAd.} NormAd \cite{rao2024normadframeworkmeasuringcultural} offers a structured collection of cultural norms and situational examples, demonstrating how abstract values materialize in everyday interactions. Unlike WVS, which provides broad statistical insights, and Wikipedia, which offers descriptive knowledge, NormAd emphasizes behavioral and contextual applications of cultural principles.

\subsection{Evaluation Dataset}\label{subsec:dataset}

We use two sets of tasks for evaluations:

\noindent\textbf{Multicultural Multilingual Offensiveness.} To assess the effectiveness of adaptation in models' behavior on downstream tasks, we evaluate the adapted models using a combination of datasets (such as OffenseEval2020, \citealt{OffenseEval2020/semeval/ZampieriNRAKMDP20}) following  \citet[see original publications or Appendix~\ref{app:test_data_statistics} for the complete list, which consists of 59 datasets]{cultureLLM/corr/abs-2402-10946, culturePark/corr/abs-2405-15145}. The test data contains a total of 68607 multilingual, culturally sensitive texts annotated for offensiveness.

\noindent\textbf{MMLU.} 
To evaluate the model's general knowledge retention capabilities after cultural adaptation, we assess each adapter's performance on factual question-answering tasks using MMLU \cite{mukherjee-etal-2024-cultural}. The MMLU dataset focuses on factual knowledge such as mathematics, biology, chemistry etc., which contains minimal cultural sensitivity. The deviations in MMLU accuracy following cultural fine-tuning would suggest unintended interference, implying the cultural adapter alters the model's underlying knowledge representations.

Using these two datasets, we enable a systematic evaluation of how effectively language models can integrate cultural perspectives into downstream tasks while preserving their factual knowledge. 

\paragraph{Dataset Balance.} Our training and evaluation sets are imbalanced across cultures and languages (see Appendix~\ref{app:statistics}; Tables~\ref{tab:wiki_stats}, \ref{tab:normad_stats}, and \ref{tb-datasets}). For example, NormAd token counts range from 102{,}705 (Arabic) to 6{,}784 (Korean), and Wikipedia tokens from 15{,}632 (English) to 3{,}662 (Spanish). To mitigate confounding, we (i) report per-culture scores alongside macro-averages, (ii) use the \diagonality{} normalization that compares each adapter against the best performer for each test culture (column-wise), and (iii) repeat all runs with three seeds and report mean~$\pm$~sd. We avoid aggressive rebalancing to preserve culture-specific signal, so some absolute gains may still correlate with data availability; we therefore interpret improvements as gains under realistic resource disparity and revisit residual skew in the Limitations.

\subsection{Models and Training}
In this work, we evaluate three variants of LLMs, including Llama-3.1-8B (base and instruction-tuned, \citealt{llm/llama/abs-2302-13971, llama3/corr/abs-2407-21783}), Gemma-2-9B (instruction-tuned, \citealt{gemma2/corr/abs-2408-00118}), and Qwen-2.5-7B  (instruction-tuned, \citealt{qwen2.5}). In our experiments, all instruction-tuned models are suffixed with ``\textbf{-IT}''. We perform LoRA adaptation~\cite{hu2021loralowrankadaptationlarge} on each model using rank-64 LoRA matrices, a batch size of 32, a learning rate of $2 \times 10^{-4}$, and six training epochs. All adapters were trained on a single 80GB A100 GPU with bf16 and gradient checkpointing. Other details on training are in Appendix \ref{app:training_reformulation}.

We evaluate three open-weight families, each with instruction-tuned variants, to ensure results generalize across distinct pretraining corpora while remaining reproducible \cite{llama3/corr/abs-2407-21783, gemma2/corr/abs-2408-00118, qwen2.5}. Restricting capacity to 7 to 9B balances multilingual coverage and compute so that PEFT fits on a single high-memory GPU. We adopt per-culture LoRA adapters instead of full fine-tuning to reduce catastrophic forgetting (tracked with MMLU; \S\ref{subsec:mmlu_interference}) and enable plug-and-play cultural control. Adapters target \texttt{q\_proj}/\texttt{v\_proj} with rank~64, dropout~0.1, batch size~32, learning rate $2{\times}10^{-4}$, and six epochs, a setting that preserved knowledge while maintaining cultural distinctiveness.

\subsection{Evaluation Metrics}
In our main paper, we evaluate each model's performance using freeform generation, assessing its ability to provide culturally relevant justifications or context. Our Appendix includes additional probability-based evaluations, using token-level likelihood scores to measure the model's confidence in classifying offensive content across cultures.
Further, we use F1 score as the primary metric for evaluating classification performance on both probability and freeform-based evaluations.

We propose a \textit{cultural distinctiveness} metric, \textbf{\textsc{C-Dist}} score, to further quantify a model's ability to preserve cultural distinctiveness. For $n$ cultures, we define a performance matrix $M \in \mathbb{R}^{n \times n}$, where $M_{i,j}$ is the F1-score when a model adapted to culture $i$ is evaluated on test data for culture $j$. We compute:

\begin{enumerate}[noitemsep, topsep=0.1pt]
    \item Extract the diagonal entries\footnote{We define ``diagonal entries" as the corresponding performance of an adapter on its corresponding culture, e.g. Korean Adapter evaluated on Korean Culture test set, hence we define this as $M_{i,i}$} $\vec{d} = [M_{i,i}]_{i=1}^n$.
    \item Normalize each $M_{i,i}$ by the maximum value in its column: $\vec{n}_i = M_{i,i} / \max_j M_{j,i}$.
    \item Average these normalized diagonal entries: 
    \begin{equation}
        D = \frac{1}{n} \sum_{i=1}^{n} \vec{n}_i.
    \end{equation}
\end{enumerate}

\begin{figure}[h]
    \centering
    \centering
    \includegraphics[width=\linewidth]{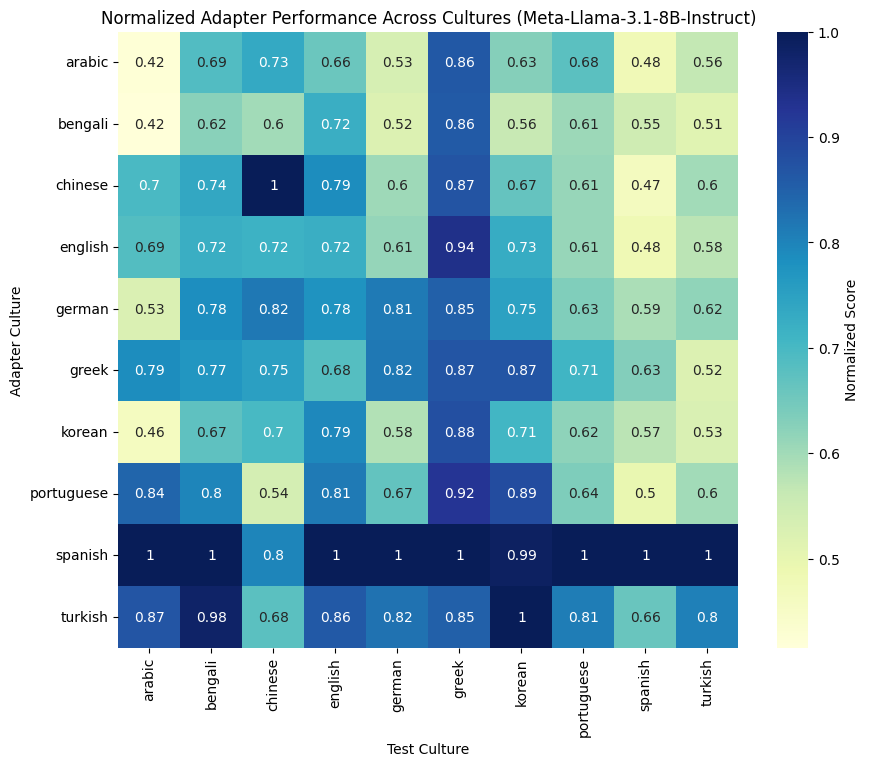}
    
    \caption{Single-culture adaptation using WVS data with Llama-3.1-8B-IT, evaluating cross-cultural offensiveness classification tasks. Minimal diagonal pattern is observed in this setting, with a \diagonality score of \textbf{0.76}.}
    \label{fig:no_diagonal}
\end{figure}

In the formula above, we normalize by column (i.e., by the test culture) since each test culture set may have different difficulty and scales. This normalization also helps identify which adapter performs best for a given culture. In an ideal scenario, the best performing adapted model for a particular culture should be based on its own culture, resulting in a \diagonality score of $1.0$. A lower score suggests interference or homogenization, as illustrated in Figure \ref {fig:no_diagonal}. This metric thus quantifies the extent to which each model preserves distinct cultural representations after adaptation.

\begin{table*}[ht!]
\centering
\footnotesize
\begin{tabular}{lrrrrrrrrrrr}
\toprule
\textbf{Model} & \textbf{ara} & \textbf{ben} & \textbf{zho} & \textbf{eng} & \textbf{deu} & \textbf{ell} & \textbf{kor} & \textbf{por} & \textbf{spa} & \textbf{tur} & \textbf{Avg.} \\
\midrule
\multicolumn{11}{c}{\textbf{Zero-Shot Prompting}} \\
\midrule
Llama-3.1-8B & 11.96 & 17.12 & 32.77 & 14.85 & 23.81 & 38.16 & 26.14 & 19.93 & 30.96 & 21.95 & 23.77 \\
Llama-3.1-8B-IT & 19.14 & 23.10 & 30.49 & 26.63 & 34.36 & 37.56 & 38.72 & 20.92 & 39.14 & 32.95 & 30.00 \\
\midrule
\multicolumn{11}{c}{\textbf{Single-Culture Adaptation - WVS}} \\
\midrule
Llama-3.1-8B & 17.22 & 22.01 & 38.28 & 19.92 & 29.30 & 36.08 & 32.65 & 20.15 & 27.93 & 28.57 & 27.21 \\
Llama-3.1-8B-IT & 19.50 & 23.51 & 32.69 & 22.35 & 34.78 & 36.98 & 37.61 & 17.75 & 25.85 & 28.78 & 27.98 \\
\bottomrule
\end{tabular}
\caption{Culture adaptation results (F1 scores) under three training scenarios: zero-shot prompting, single-culture adaptation training on Llama-3.1-8B models using WVS training data. The adaptation is evaluated using a multilingual offensiveness dataset (\S\ref{subsec:dataset}) reported with averaged F1 scores. 
}
\label{tab:combined-baseline}
\end{table*}

\section{Adaptation with WVS:  Findings and Observed Interferences}
\label{sec:baselines}

In this section, we focus on Llama-3.1-8B models (both base and instruction-tuned) to establish a clear understanding of their performance and the impact of adaptation using WVS data, including cultural and knowledge-based interferences.

\subsection{Performance Gains Driven by Enhanced Instruction Following}
\label{subsec:inconsistency}

\noindent\textbf{General Observations.} Table~\ref{tab:combined-baseline} compares the approaches for downstream tasks using Llama-3.1-8B models: (i) zero-shot prompting, (ii) single-culture adaptation. Our results show that training using WVS is more effective in improving downstream tasks for the base model when using the single-culture adaptation strategy. Particularly, WVS training is beneficial for underrepresented cultures such as \texttt{ara} and \texttt{kor}. Surprisingly, this positive effect is not seen in the instruction-tuned model, which instead shows a decline in performance.

\begin{table}[h]
\centering
\footnotesize
\begin{tabular}{llr}
\toprule
\textbf{Methods} &  & \textbf{Invalid (\%)} \\
\midrule
Llama-3.1-8B & Zero-Shot & 20.12 \\
& Single-Culture-WVS & 14.68 \\
\midrule
Llama-3.1-8B-IT & Zero-Shot & 21.20 \\
& Single-Culture-WVS & 10.82 \\
\midrule
Gemma & Zero-Shot & 11.75 \\
& Single-Culture-WVS & 0 \\
\midrule
Qwen & Zero-Shot & 6.8 \\
& Single-Culture-WVS & 0 \\
\bottomrule
\end{tabular}
\caption{Comparison of invalid response rates across different models and scenarios. The Invalid Ratio represents the percentage of responses flagged as invalid across all culture test sets. We provide the complete invalid ratio table in Appendix~\ref{app:full_invalid_ratio}.}
\label{tab:invalid-responses-m}
\end{table}

\noindent\textbf{Performance Gain by Better Instruction Following.} 
To understand why the instruction-tuned model did not benefit from training with WVS, we analyze its downstream task predictions by examining the ratio of invalid responses\footnote{An invalid response contains nonsensical outputs, fails to follow instructions or lacks a meaningful or relevant answer to the prompt. Appendix~\ref{apptab:invalid-responses} shows example responses.} before and after adaptation in Table~\ref{tab:invalid-responses-m} (completed results in Appendix \ref{app:invalid_rate}). Compared to zero-shot prompting, both the base model and instruction-tuned model have significantly improved invalid response ratios after adaptation. This suggests that WVS fine-tuning enhances the model's general instruction-following ability but does not necessarily improve its understanding of cultural values. 

The high zero-shot invalid response ratio in models shows that achieving strong performance on relevant tasks requires improvements in \emph{both} instruction-following ability and cultural value understanding.

\subsection{Observed Cultural Interference Across Models}
\label{subsec:cultural_interference}
To further investigate the effect of adaptation, we examine the single-culture adaptation results in a cross-cultural setting (i.e., training on one culture and evaluating on others). Ideally, performance should be highest when the adaptation matches the test culture, forming a diagonal pattern in a heatmap of cross-cultural evaluations. However, as shown in Figure~\ref{fig:no_diagonal}, no such diagonal is observed for the instruction-tuned Llama model (with a similar pattern seen for the base model in Figure~\ref{fig:no_diagonal_app} in the Appendix). The cross-cultural improvements show no clear trends, and all adapters enhance performance on the Spanish test data in Figure~\ref{fig:no_diagonal}. The \diagonality score (introduced in \S\ref{sec:experimental_setup}) remains below $0.80$ for both models.

The results further suggest that WVS is not necessarily the best data source for improving cultural values, as the adapted models fail to preserve their own culture's perspectives, leading to compromised cross-cultural result improvements (i.e., \emph{cultural interference}).

\begin{table}[h!]
\centering
\small
\begin{tabular}{l lcc}
\toprule
\textbf{Model} & \textbf{Culture} & \textbf{Std.} & \textbf{Transl.} \\
\midrule
\multirow{10}{*}\textbf{Llama-3.1-8B}
    & ara & 32.24 & 32.83 \\
    & ben & 48.67 & 51.81 \\
    & zho & 38.21 & 41.08 \\
    & eng & 23.00 & 29.58 \\
    & deu & 33.55 & 39.68 \\
    & ell & 30.75 & 31.55 \\
    & kor & 27.59 & 27.57 \\
    & por & 46.41 & 28.77 \\
    & spa & 35.53 & 35.27 \\
    & tur & 19.74 & 18.02 \\
    & Avg. & 33.57 & 33.62 \\

\midrule
\multirow{10}{*}\textbf{Llama-3.1-8B-IT}     
    & ara & 41.99 & 37.81 \\
    & ben & 45.45 & 42.77 \\
    & zho & 41.35 & 46.28 \\
    & eng & 42.81 & 49.18 \\
    & deu & 40.40 & 41.92 \\
    & ell & 46.05 & 36.34 \\
    & kor & 41.80 & 44.63 \\
    & por & 40.11 & 38.08 \\
    & spa & 43.77 & 38.60 \\
    & tur & 43.93 & 40.46 \\
    & Avg. & 42.78 & 41.61 \\
\bottomrule
\end{tabular}
\caption{MMLU evaluation after single-culture adaptation with WVS data (F1 Score \%). Performance variation is evident across cultural adapters, with observed factual knowledge retention and potential cultural biases. The zero-shot performance is \textbf{35.05} for Llama-3.1-8B and \textbf{45.38} for Llama-3.1-8B-IT.}
\label{tab:llama-results}
\end{table}

\begin{figure*}[ht!]
\centering
\begin{subfigure}[b]{0.32\textwidth}
    \centering
    \includegraphics[width=\textwidth]{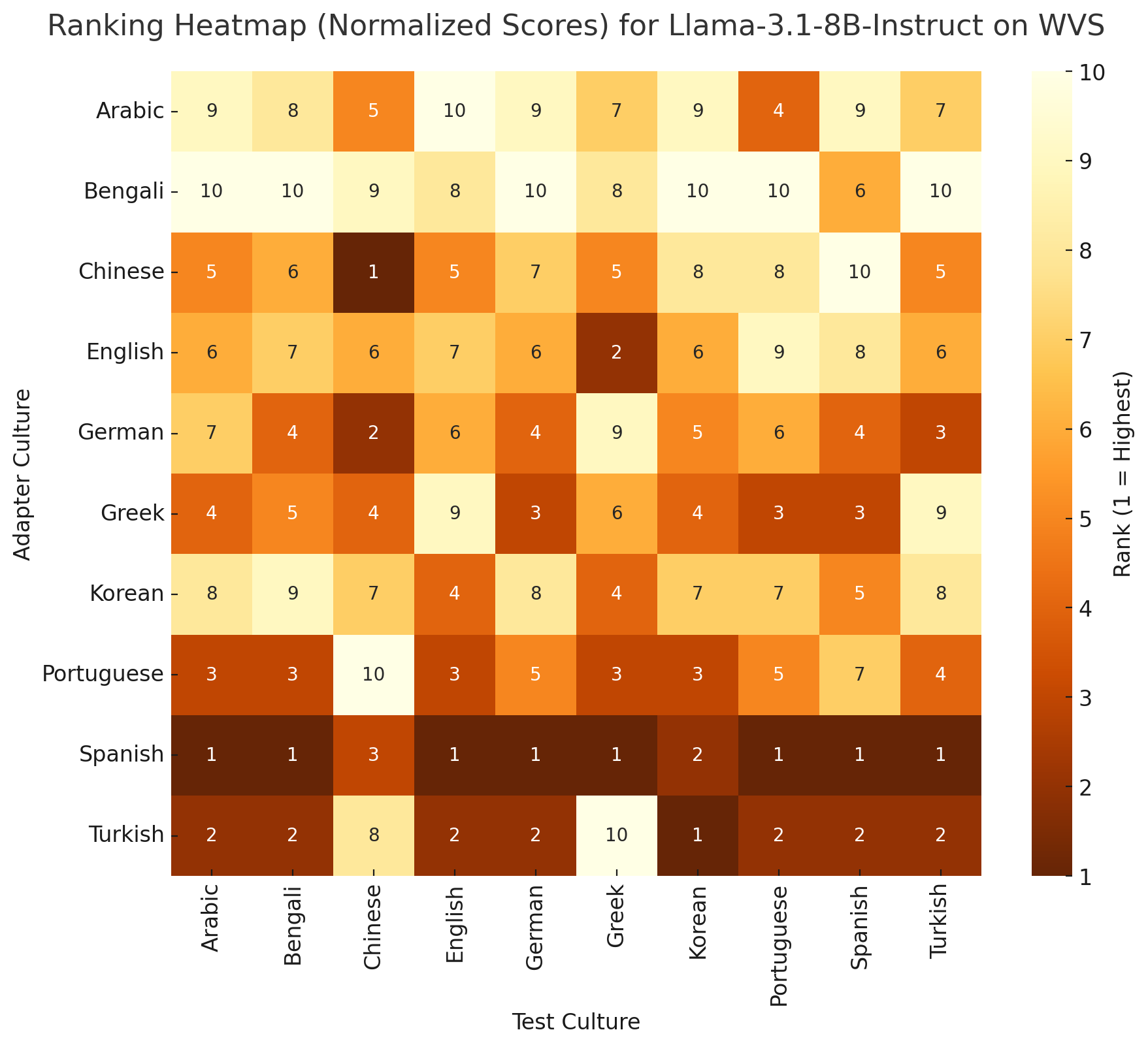}
    \caption{\textbf{WVS}}
\end{subfigure}
\hfill
\begin{subfigure}[b]{0.32\textwidth}
    \centering
    \includegraphics[width=\textwidth]{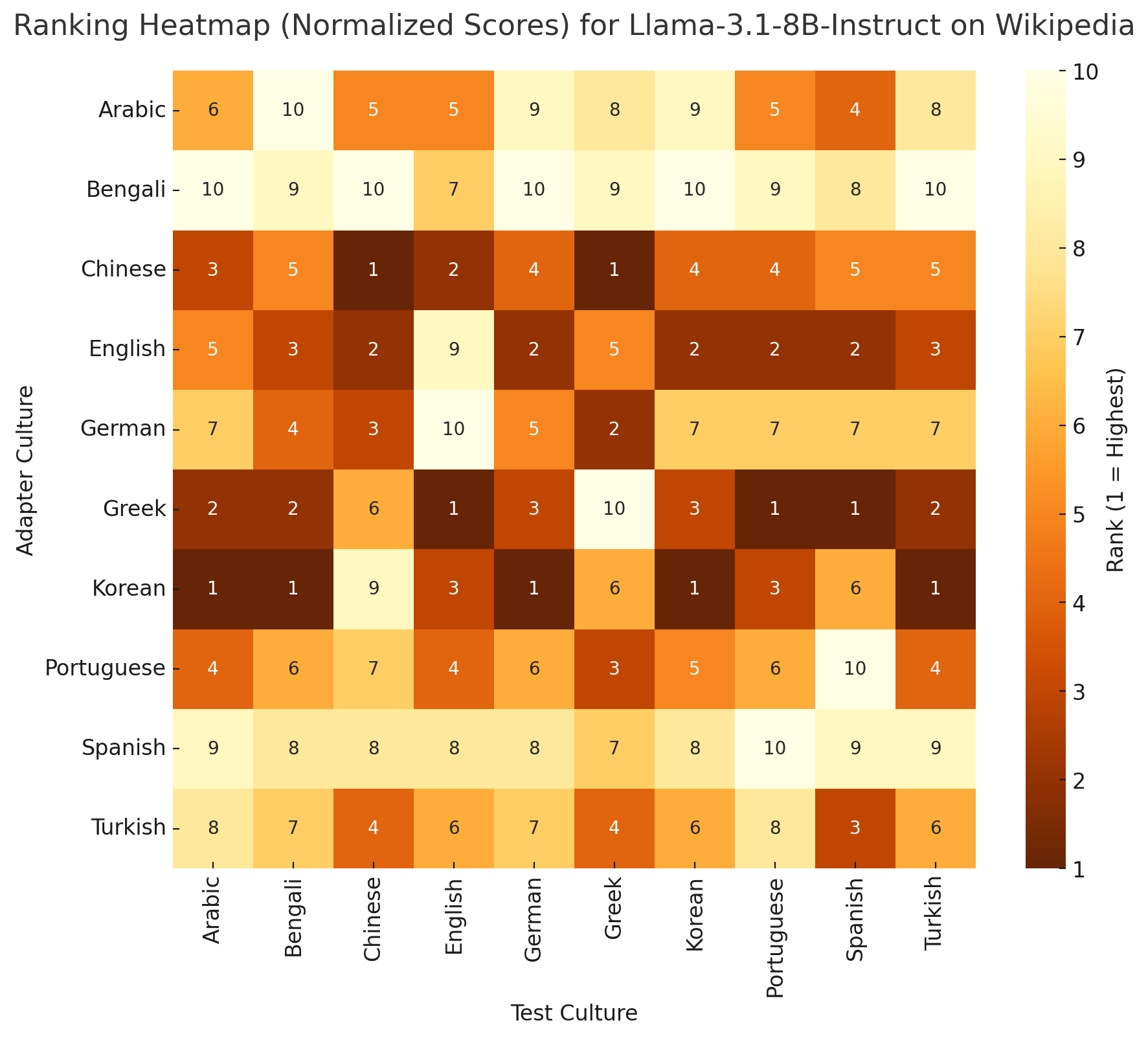}
    \caption{\textbf{WVS+Wiki}}
\end{subfigure}
\hfill
\begin{subfigure}[b]{0.32\textwidth}
    \centering
    \includegraphics[width=\textwidth]{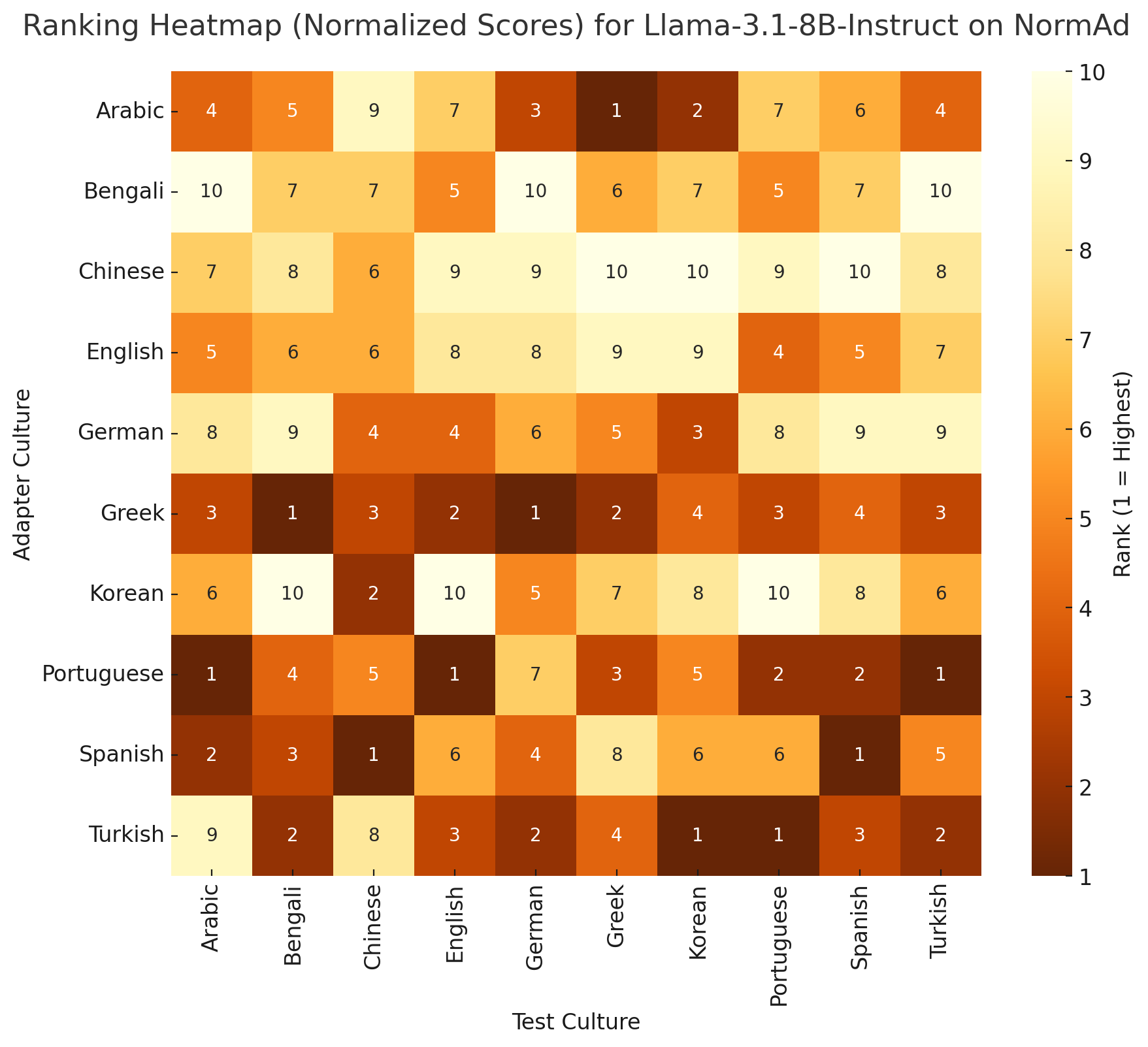}
    \caption{\textbf{WVS+NormAd}}
\end{subfigure}
\caption{Heatmaps of culture-specific classification performance (Llama-3.1-8B-IT) based on the ranks of the adaptation results. Darker diagonal elements indicate stronger cultural distinctiveness and better \diagonality scores.}
\label{fig:three_clusters}
\end{figure*}

\subsection{Factual Knowledge Interference}
\label{subsec:mmlu_interference}

Fine-tuning improves cultural alignment but may unintentionally impact factual knowledge \cite{mukherjee-etal-2024-cultural}. Ideally, cultural value adaptation should not affect objective QA performance.

Table~\ref{tab:llama-results} presents the results of single-culture adaptation on MMLU. Both Llama-3.1-8B and Llama-3.1-8B-IT exhibit significant variability when trained under two conditions: standard (using English WVS data) and translated (WVS values in their respective languages). Additionally, the base model shows a decline in performance compared to zero-shot prompting, while the instruction-tuned model shows performance improvements.

These fluctuations in the results show that adapting to WVS data can change factual knowledge accuracy, depending on language and dataset characteristics. Furthermore, the inconsistencies in probability-based scoring (Appendix~\ref{tab:mmlu_wvs_proba}) also strengthen the observation of \emph{factual knowledge interference}. This underscores the challenge of balancing cultural distinctiveness with factual integrity with the appropriate training data.

\begin{table*}[h!]
\footnotesize
\centering
\begin{tabular}{l l c | c c}
\toprule
\textbf{Model} & \textbf{Data} & \textbf{\diagonality} & \textbf{F1 Cult. (\%)} & \textbf{F1 MMLU (\%)} \\
\midrule
\multirow{6}{*}{Llama-3.1-8B-IT} 
  & WVS              & $0.76 \pm 0.012$  & $29.61 \pm 1.91$ & $42.78 \pm 0.84$ \\
  & Wiki             & $0.81 \pm 0.009$  & $35.39 \pm 2.77$ & $26.33 \pm 2.05$ \\  
  & NormAd           & $0.85 \pm 0.014$  & $38.42 \pm 2.92$ & $19.63 \pm 1.38$ \\
  & WVS+Wiki         & $0.78 \pm 0.010$  & $31.19 \pm 1.80$ & $49.02 \pm 0.88$ \\
  & WVS+NormAd       & $\mathbf{0.89 \pm 0.011}$  & $\mathbf{40.94 \pm 0.98}$ & $\mathbf{50.43 \pm 0.93}$ \\
  & WVS+Wiki+NormAd  & $0.76 \pm 0.013$  & $38.21 \pm 0.85$ & $52.61 \pm 0.97$ \\
\midrule
\multirow{6}{*}{Gemma-2-9B-IT} 
  & WVS              & $0.81 \pm 0.011$  & $39.22 \pm 1.66$ & $45.31 \pm 0.78$ \\
  & Wiki             & $0.83 \pm 0.010$  & $36.67 \pm 2.61$ & $8.23 \pm 1.92$ \\  
  & NormAd           & $0.79 \pm 0.013$  & $37.10 \pm 2.64$ & $8.07 \pm 1.96$ \\
  & WVS+Wiki         & $0.80 \pm 0.012$  & $37.25 \pm 1.60$ & $47.05 \pm 0.82$ \\
  & WVS+NormAd       & $\mathbf{0.83 \pm 0.012}$  & $\mathbf{40.01 \pm 0.69}$ & $\mathbf{55.19 \pm 0.86}$ \\
  & WVS+Wiki+NormAd  & $0.73 \pm 0.018$  & $37.90 \pm 0.83$ & $64.94 \pm 1.01$ \\
\midrule
\multirow{6}{*}{Qwen2.5-7B-IT} 
  & WVS              & $0.92 \pm 0.006$  & $48.05 \pm 1.41$ & $68.32 \pm 0.56$ \\
  & Wiki             & $0.89 \pm 0.008$  & $44.21 \pm 2.38$ & $58.32 \pm 1.62$ \\  
  & NormAd           & $0.91 \pm 0.007$  & $48.31 \pm 2.42$ & $65.57 \pm 0.58$ \\
  & WVS+Wiki         & $0.90 \pm 0.007$  & $46.00 \pm 1.39$ & $68.22 \pm 1.57$ \\
  & WVS+NormAd       & $\mathbf{0.94 \pm 0.006}$  & $47.67 \pm 0.40$ & $67.51 \pm 0.57$ \\
  & WVS+Wiki+NormAd  & $0.86 \pm 0.010$  & $44.13 \pm 0.37$ & $67.33 \pm 0.98$ \\
\bottomrule
\end{tabular}
\caption{Mean $\pm$ standard deviation across 3 runs with different seeds. \textbf{NormAd} drives the largest gains in cultural distinctiveness (\diagonality) and cultural-task F1, especially when paired with WVS (Llama/Gemma). \textbf{Wikipedia} alone yields smaller C-DIST gains but, when combined with WVS, stabilizes or improves MMLU (knowledge retention). Variability (sd) is modest for C-DIST and F1, and slightly higher for narrative-only MMLU rows, consistent with greater instability when training without WVS grounding.}
\label{tab:diagonality-results}
\end{table*}

\section{Adaptation with Additional Narratives}
\label{sec:enhanced_adaptation}

While WVS-based training provides a useful starting point for cultural value adaptation, our results in Section~\ref{sec:baselines} show that it rarely produces strong diagonal patterns, suggesting limited cultural specialization. This raises a key question: \emph{what types of additional data can strengthen adaptation while preserving cultural distinctiveness?}

Decades of work in social psychology highlight a long standing gap between what people \emph{report} in surveys and how they \emph{behave} in practice \cite[inter alia]{gross1975attitude, fazio1981direct}. This raises concern on the sufficiency of self-reported values such as those in WVS for improving tasks that depend on culturally grounded behavior (\S\ref{subsec:inconsistency}). To bridge this gap, we add WVS with two complementary narrative sources: Wikipedia and NormAd. Unlike survey responses, these provide more objective, context-rich source of cultural norms, institutions, and everyday practices. We hypothesize that incorporating such \emph{narratives of culture} helps models learn both abstract values and their behavior.

Here, we focus our evaluation on instruction-tuned models to better reflect real-world use and extend it beyond Llama to include Gemma and Qwen, demonstrating the generality.\\

\subsection{Complementary roles of Wikipedia and NormAd}
\label{subsec:wiki_vs_normad}
Beyond surveys, our additional sources contribute different kinds of supervision. Wikipedia introduces largely declarative, encyclopedic context about institutions, historical trajectories, and everyday practices. Because this material is descriptive and source-based, it tends to regularize the model toward factual consistency and mitigate catastrophic forgetting during cultural finetuning. By contrast, NormAd supplies procedural, scenario-based norms-roles, obligations, and sanctions expressed in concrete situations, so the training signal is closer to how cultural rules are applied in behavior-sensitive settings.

Empirically (Table~\ref{tab:diagonality-results}), combining NormAd on top of WVS (\textit{WVS+NormAd}) gives the largest gains in cultural distinctiveness and culture-task F1. For example, on Llama-3.1-8B-IT we see \diagonality $0.76{\to}0.89$ and F1-Cult.\ $29.61{\to}40.94$, along with a clearer diagonal in Figure~\ref{fig:three_clusters}. Adding Wikipedia to WVS (\textit{WVS+Wiki}) usually keeps or improves MMLU while maintaining cultural separation, and using all three together often gives the best MMLU for Llama/Gemma.

We observe that this happens due to NormAd contains concrete, situation-based rules, so the model learns how to apply norms in decisions. Wikipedia provides broad factual background, which helps keep general knowledge stable and offsets drift from training only on surveys (WVS). As a result, narrative-only fine-tuning can make knowledge less stable (Appendix~\ref{app:full_invalid_ratio}), and WVS-only adapters tend to blur across cultures (\S\ref{subsec:cultural_interference}). Mixing \textit{WVS+NormAd} restores cultural separation, and adding Wikipedia yields the most reliable balance overall.

\begin{figure*}[ht!]
    \centering
    \subfloat[UMAP KDE on Wikipedia]{\includegraphics[width=0.40\linewidth]{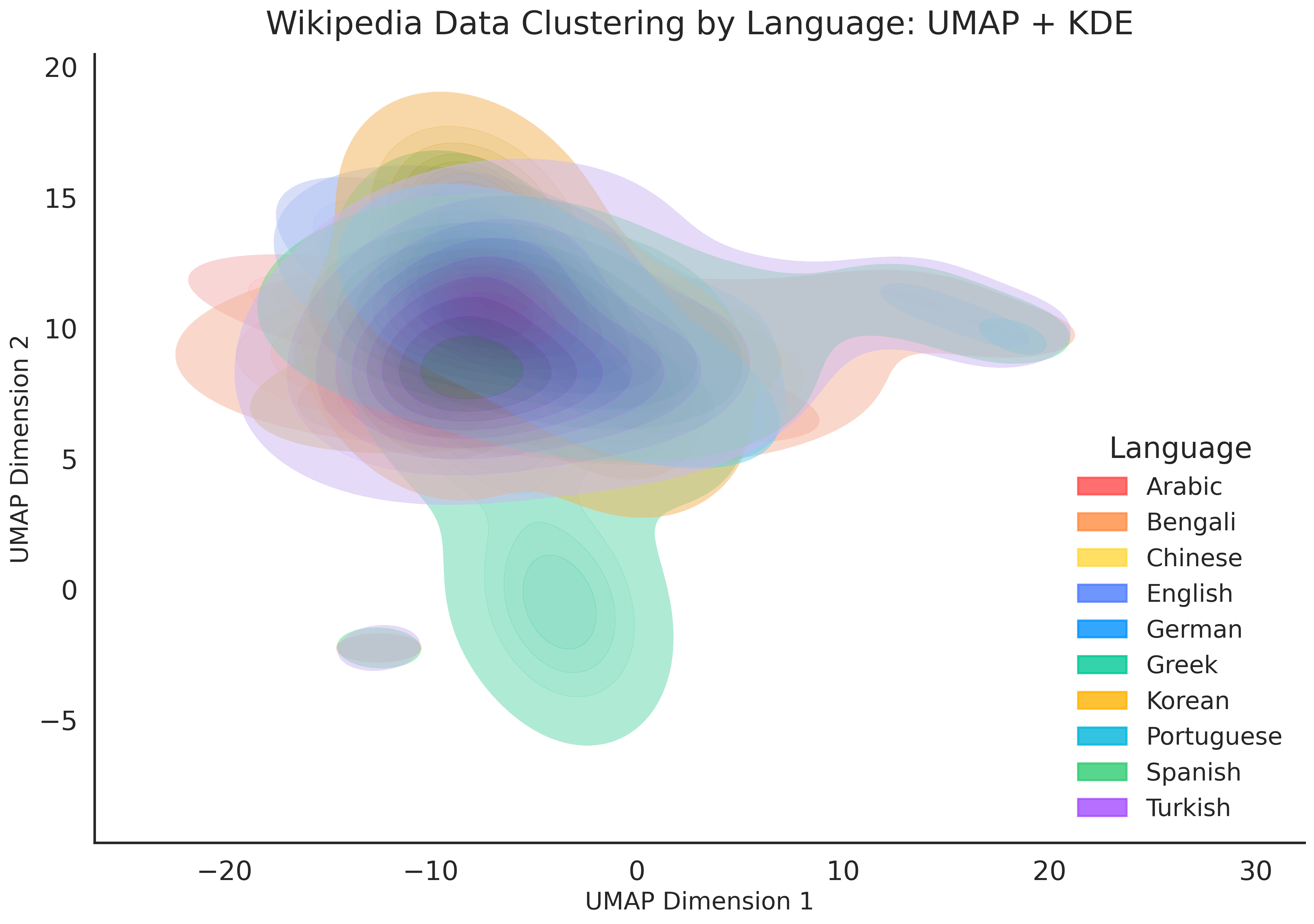}}
    \subfloat[UMAP KDE on NormAd]{\includegraphics[width=0.40\linewidth]{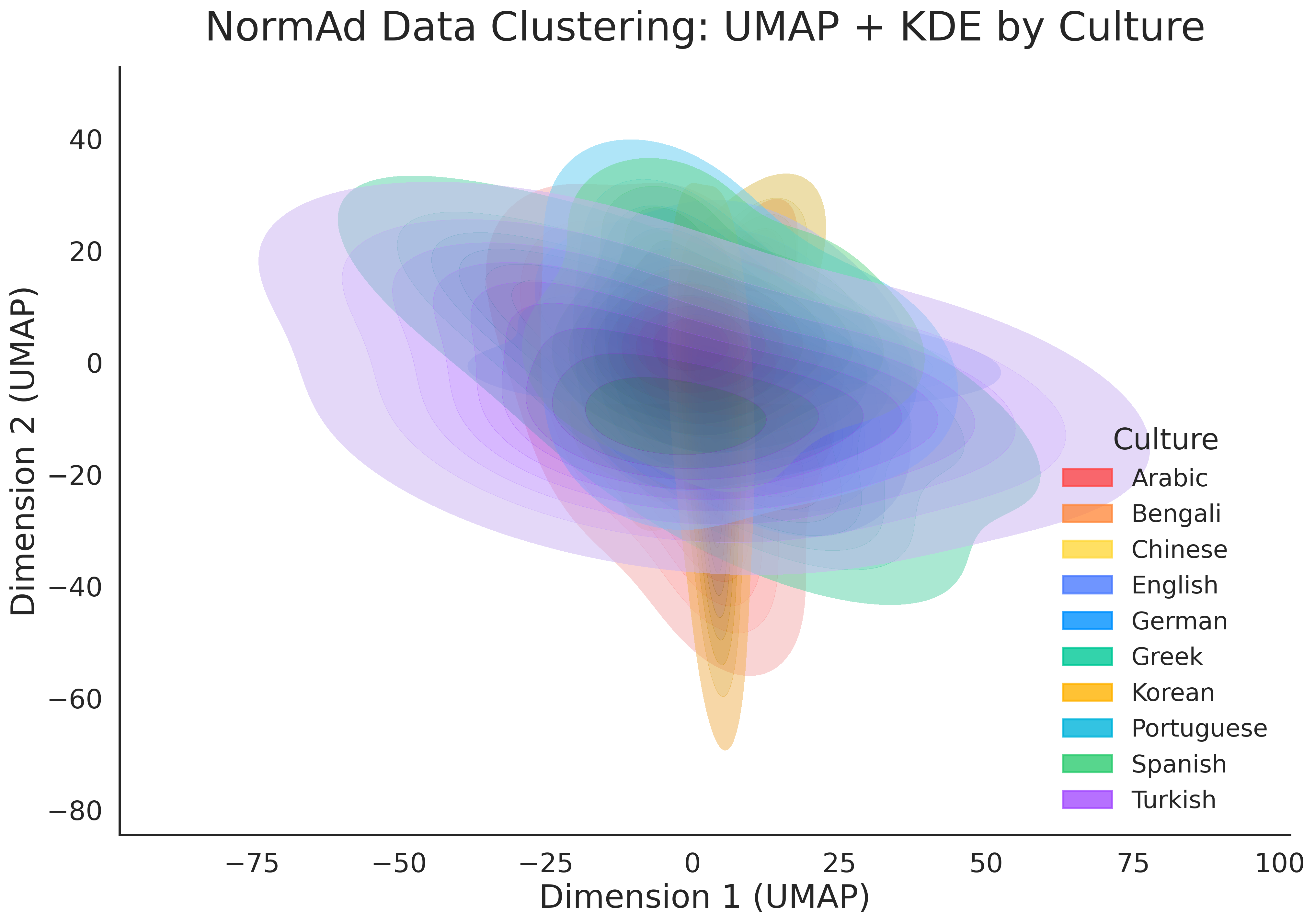}}
    \caption{Kernel Density Estimation (KDE) plots of UMAP embeddings using LaBSE \cite{feng2022languageagnosticbertsentenceembedding} for Wikipedia and NormAd datasets. These visualizations show the density distributions of the data in the reduced-dimensional space.}
    \label{fig:umap_kde}
\end{figure*}

\noindent\textbf{Improved \diagonality with Wikipedia and NormAd.}
The addition of Wikipedia and NormAd often enhances cultural distinctiveness, with improvements statistically significant in most cases (see Table \ref{tab:diagonality-results}). Table~\ref{tab:diagonality-results} shows that integrating these datasets consistently improves \diagonality scores across all three models, indicating more culturally distinct behavior. For instance, Llama-3.1-8B-IT's \diagonality improves from $0.76$ (WVS-only) to $0.89$ (WVS+NormAd). Figure~\ref{fig:three_clusters} illustrates this shift, as the heatmaps become more diagonal and show reduced cross-cultural interference. \emph{Incorporating additional cultural narratives retains cultural distinctiveness.}

\noindent\textbf{Improve over WVS alone in many cases.} The addition of Wikipedia and NormAd data leads to notable gains in offensive classifications compared to training with WVS data alone. For instance, Llama-3.1-8B-IT's performance on the offensiveness classification tasks (denoted as \textbf{F1 Cult.} in Table \ref{tab:diagonality-results}) rises from 29.61\% (WVS-only) to 40.94\% (WVS+NormAd), reflecting the value of richer, context-laden cultural information. However, Gemma-2-9B-IT and Qwen2.5-7B-IT see a marginal change in F1 Cult., when WVS is augmented with NormAd. This highlights that while Llama-3.1-8B-IT showed clear benefits on this downstream task from narrative augmentation, the effect on tasks is model-dependent.

For MMLU, combining WVS with Wikipedia and NormAd (WVS+Wiki+NormAd) yields the best results for both Llama-3.1-8B-IT and Gemma-2-9B-IT. However, our results show anomalies, indicating the ongoing challenge of achieving robust cultural adaptation without compromising general knowledge retention. Further, the general trend indicates that context-rich data, \emph{when added to WVS}, effectively helps offset the knowledge interference introduced by survey data alone. Overall, our findings suggest that \emph{curated narratives are crucial for retaining the model’s foundational understanding of cultural knowledge during adaptation.}

\section{Further Analysis}
Our empirical results suggest that adding objective cultural descriptions and context-specific examples improves cultural distinctiveness and performance on downstream tasks. We also found that pairing WVS with scenario-based narratives (WVS+NormAd; optionally +Wiki) is a strong default when aiming for culturally distinct behavior without sacrificing factual knowledge. In this section, we analyze the data further to understand why.

\noindent\textbf{Overlapping Embeddings versus Distinct Adaptations.} We first embed each data source using LaBSE (\citealt{feng2022languageagnosticbertsentenceembedding}, a multilingual embedding model that compresses texts into a shared semantic space), then project the embedding with kernel density estimation (KDE). The results for WVS, Wikipedia and NormAd are shown in Figure \ref{fig:wvs_clustering} and Figure~\ref{fig:umap_kde} respectively. It is interesting to note that there is no distinct separation between cultures within a dataset. This suggests that semantic differences in the data are not the primary factor influencing downstream differences after training.

This discrepancy likely occurs because Wikipedia and NormAd differ in \emph{how} they encode cultural details, even if their embeddings are not sharply separated (see Table \ref{apptab:data_samples} in Appendix for data examples). Wikipedia provides broad encyclopedic summaries, covering historical contexts and traditions, while NormAd provides scenario-specific norms that directly inform cultural behaviors (e.g., respecting elders in formal gatherings).
These nuanced differences at the domain level do not necessarily create distinct embedding clusters. Nevertheless, the descriptive, scenario-based NormAd dataset enhances fine-tuning by providing more targeted cultural cues. As a result, the model can better isolate culture-specific behaviors, yielding higher \diagonality scores.

\subsection{Summary of Findings}
\label{subsec:summary_findings}
Fine-tuning on WVS data alone is ineffective for cultural value adaptation, as shown by low \diagonality scores, weaker downstream task performance, and reduced factual knowledge retention. While overall performance may vary across tasks, augmenting survey data with more descriptive sources enables a model to \emph{retain cultural distinctiveness} and \emph{retain factual knowledge} better. Combining WVS survey data with NormAd situational norms consistently yields clearer cultural separation, as evidenced by improved \diagonality score (Table~\ref{tab:diagonality-results}). 
Wikipedia data offers moderate gains through structured knowledge, but NormAd's scenario-based behavioral cues drive stronger cultural differentiation when paired with WVS.

Our findings suggest that combining scenario-based narratives (e.g., NormAd) with survey patterns (WVS) better preserve cultural distinctiveness and should be investigated further.

\section{Related Work}
\label{sec:related_work}

\noindent\textbf{General Adaptation to Cultural Values.} Several existing work approaches cultural value adaptations in LLMs through prompting \cite{nvm/anthroprompt/acl/abs-2402-13231, thanksgiving/acl/wang-etal-2024-countries, pnas/tao2024cultural}, continual pre-training on diverse multilingual data \cite{thanksgiving/acl/wang-etal-2024-countries, echoes/acl/choenni-etal-2024-echoes} or direct tuning on survey data or synthetic data based on survey \cite{cultureLLM/corr/abs-2402-10946, DBLP:journals/corr/abs-2410-12971,  culturePark/corr/abs-2405-15145}. In particular, the basis of our investigation, CultureLLM \cite{cultureLLM/corr/abs-2402-10946}, employs semantically augmented data from the World Values Survey (WVS) to represent the average opinion of a culture. In this paper, we extend the investigation using descriptive cultural principles and provide a comprehensive analysis.  

Recent research also explored value prediction with In-Context Learning (ICL)-based adaptation methods \cite{selfalign/icl/abs-2408-16482, IndieValueReasoner/corr/abs-2410-03868, myung2025blendbenchmarkllmseveryday}. Particularly, \citet{IndieValueReasoner/corr/abs-2410-03868} showed a mild inconsistency when models adapted using individual data from one continent were evaluated using data from another (e.g., training data for other continents generally improves alignment to Oceania people). Similarly, \citet{beck-etal-2024-sensitivity} introduced sensitivity tests to probe how models respond to culturally charged questions, revealing both strengths and blind spots. These evaluation-oriented studies complement adaptation methods by highlighting where misalignments occur and by motivating the development of more nuanced cultural adaptation strategies. While related to our work, we focus on the impact at the country level rather than the broader continent level. 

\noindent\textbf{Pluralistic Alignment.}
Related to cultural value adaptation, recent studies advocate for pluralistic alignment \cite{pluralistic/icml/SorensenMFGMRYJ24}, wherein a model should reflect the values of multiple stakeholders or sub-groups. \citet{modularpluralism/emnlp/feng-etal-2024-modular} proposed a modular pluralistic alignment method, which primarily focuses on integrating diverse opinions. This research direction differs from typical existing cultural value adaptation work, which mainly focuses on reflecting the averaged value of a culture \cite[inter alia]{cultureLLM/corr/abs-2402-10946, culturePark/corr/abs-2405-15145, pnas/tao2024cultural, nvm/anthroprompt/acl/abs-2402-13231, echoes/acl/choenni-etal-2024-echoes}. 

\noindent\textbf{Cultural Inconsistencies in LLMs.} Recent work highlights the challenges LLMs face in maintaining consistent cultural values across different linguistic and social contexts \cite{adilazuarda-etal-2024-towards, beck-etal-2024-sensitivity}. One of the reasons why these inconsistencies arise is due to biases in training data \cite{mihalcea2024aiweirdwayai, sorensen-etal-2022-information}, which often prioritize Western or English-centric perspectives, leading to misalignment when applied to non-WEIRD cultures \cite{mihalcea2024aiweirdwayai}. Additionally, \citet{mukherjee-etal-2024-cultural}, shows that even the current LLMs are prone to a slight cultural and noncultural perturbation even on factual questions such as MMLU. This work builds upon the findings on how existing adaptation strategies address cultural disparities in downstream tasks.

\section{Conclusion}
\label{sec:conclusion}
In this paper, we investigated the limitations of using World Values Survey (WVS) data for cultural value adaptation in LLMs and explored the potential of augmenting it with scenario-based cultural narratives. Our findings reveal that relying solely on WVS can lead to homogenized cultural representations and interfere with factual knowledge. We demonstrate that incorporating encyclopedic (Wikipedia) and scenario-based (NormAd) narratives, particularly the latter, significantly enhances the cultural distinctiveness of adapted models.

While some variations in results were observed, we found that the augmentation could still improve nuanced cultural representations and preserve factual knowledge in models. Our findings reveal a complex trade-off between cultural distinctiveness, task performance, and knowledge retention, highlighting the need for further research on optimal data combinations and adaptation strategies to balance these competing objectives.

\section*{Limitations}
In this work, we focus on a select set of data as the source data for adaptation, including the World Values Survey (WVS), Wikipedia, and NormAd. While these datasets offer diverse cultural signals, they each come with inherent biases. For instance, WVS could be subject to self-reporting biases, Wikipedia reflects editorial biases, and NormAd consists of curated examples that may not fully represent all cultural variations.

Furthermore, our evaluation is limited to selected culturally sensitive tasks, which may not fully capture the broader range of tasks needed to assess how cultural value adaptation influences behavior. However, such an investigation requires careful task design and is beyond the scope of this work.

\section*{Ethics Statement}

Our work aims to enhance cultural value adaptations in NLP systems while carefully considering potential societal impacts. While this research may help reduce Western-centric bias and improve offensive content classification by incorporating diverse cultural values, we acknowledge the risks of potential misuse, including cultural stereotyping and demographic profiling. We emphasize that our findings should be applied thoughtfully, with continuous consideration of cultural context, while being careful not to anthropomorphize LLMs by attributing to them true cultural understanding or awareness. Additionally, we encourage future research to develop more nuanced methodologies and evaluation frameworks that better represent cultural diversity in NLP systems.

\section*{Acknowledgment}

This work has been supported by the LOEWE Distinguished Chair ``Ubiquitous Knowledge Processing'', LOEWE initiative, Hesse, Germany (Grant Number: LOEWE/4a//519/05/00.002(0002)/81). This research work has also been funded by the German Federal Ministry of Research, Technology and Space and the Hessian Ministry of Higher Education, Research, Science and the Arts within their joint support of the National Research Center for Applied Cybersecurity ATHENE. We thank Yongxin Huang, German Ortiz for their feedback on a draft of this paper.  

\bibliography{custom}
\bibliographystyle{acl_natbib}

\appendix
\clearpage

\onecolumn
\section{Data Characteristics}
\label{app:data_distributions}

\subsection{Additional KDE Plots}

\begin{figure*}[h!]
    \centering
    \subfloat[t-SNE KDE on Wikipedia]{\includegraphics[width=0.24\linewidth]{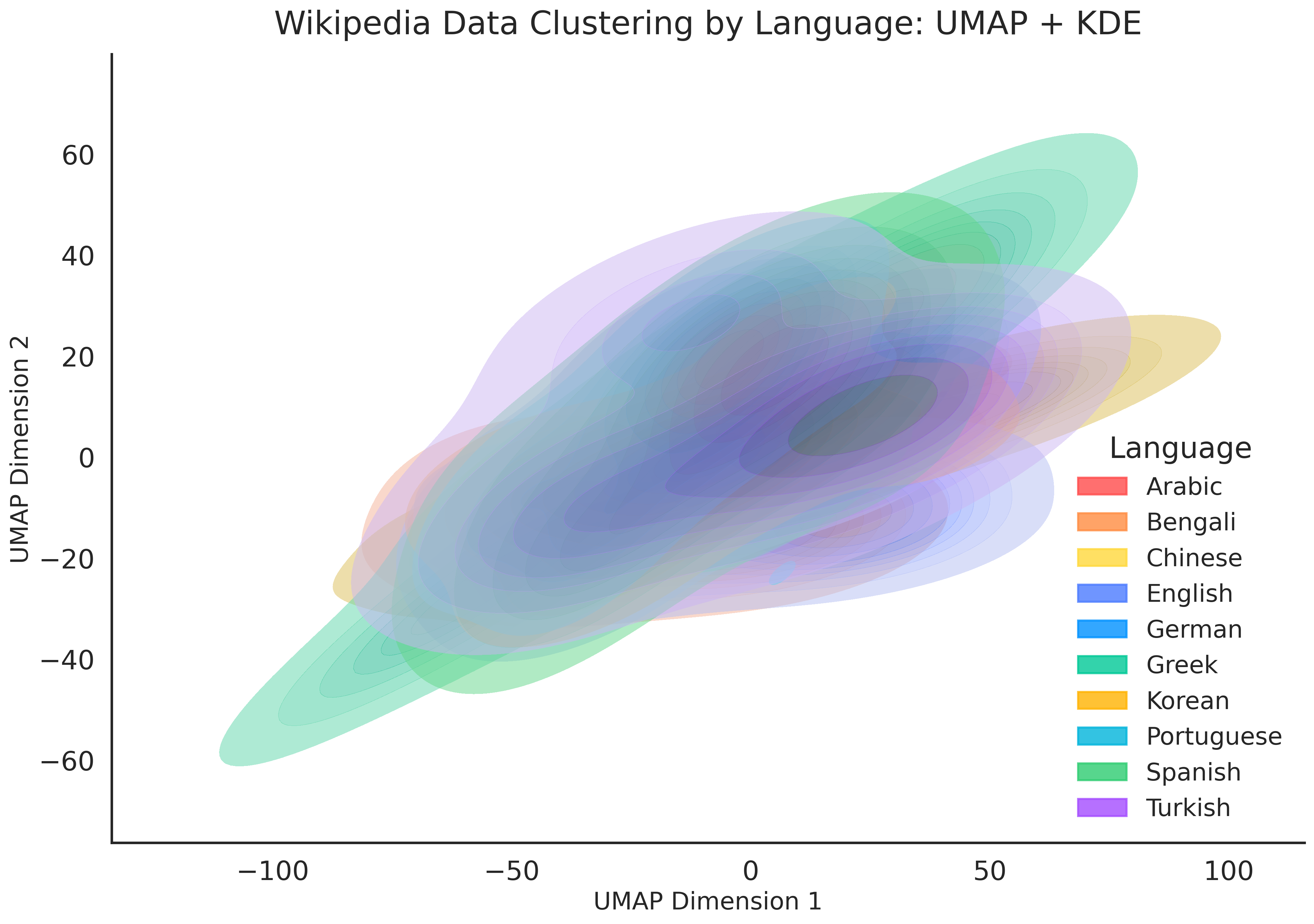}}\hfill
    \subfloat[t-SNE KDE on NormAd]{\includegraphics[width=0.24\linewidth]{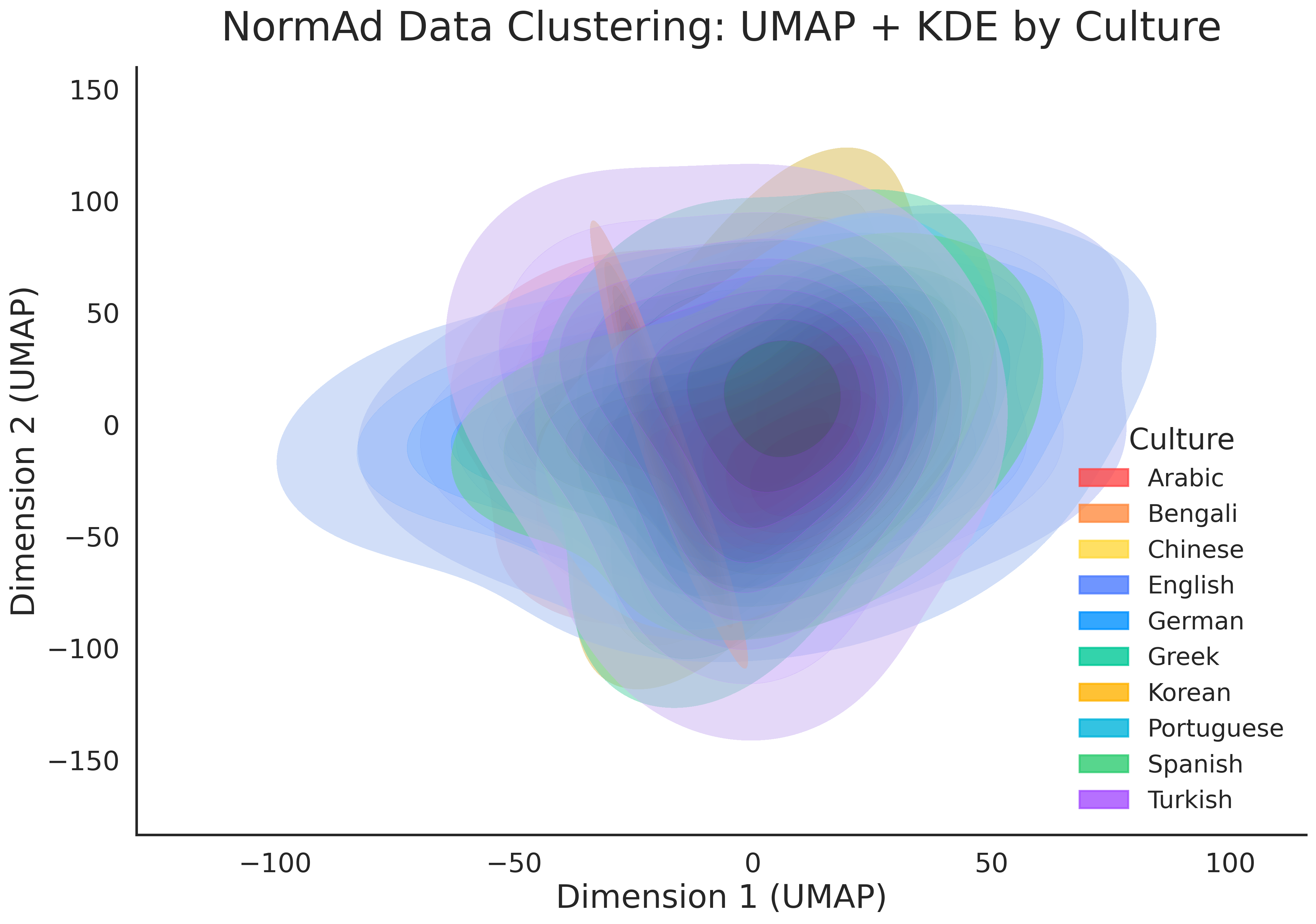}}\hfill
    \subfloat[UMAP KDE on Wikipedia]{\includegraphics[width=0.24\linewidth]{assets/wikipedia_umap_kde.png}}\hfill
    \subfloat[UMAP KDE on NormAd]{\includegraphics[width=0.24\linewidth]{assets/normad_umap_kde.png}}
    \caption{Kernel Density Estimation (KDE) plots using t-SNE and UMAP projections for Wikipedia and NormAd datasets. Although projection methods vary, none of the embeddings are distinctly separable by culture, indicating shared semantic similarities of data.}
    \label{appfig:umap_tsne_kde}
\end{figure*}

\subsection{Samples of WVS, Wiki, and NormAd Data}
Table~\ref{apptab:data_samples} presents a comparison of social values across different cultures by showcasing sample data from the World Values Survey (WVS), Wikipedia, and the NormAd dataset.

\vspace{-5pt}
\begin{table*}[h!]
    \centering
    \footnotesize
    \begin{tabular}{p{3cm} p{5cm} p{6cm}}
        \toprule
        \textbf{WVS} & \textbf{Wikipedia} & \textbf{NormAd} \\
        \midrule
        {"topic": "SOCIAL VALUES", "q\_id": "27", "q\_content": "One of my main goals in life has been to make my parents proud", "option": "1. Strongly agree 2. agree 3. Disagree 4. Strongly disagree"} & Arab culture is the culture of the Arabs, from the Atlantic Ocean in the west to the Arabian Sea in the east, in a region of the Middle East and North Africa known as the Arab world. The various religions the Arabs have adopted throughout their history and the various empires and kingdoms that have ruled and took lead of the civilization have contributed to the ethnogenesis and formation of modern Arab culture. & (Egypt - Background) \newline \textbf{Basic Etiquette} \newline - It is considered impolite to point the toe, heel or any part of the foot toward another person. Showing the sole of one's shoe is also impolite. \newline - Modest dress and presentation is highly valued in Egyptian culture. \newline - Greetings often occur before any form of social interaction. For example, a person joining a group is expected to greet all those present. \newline - Generally, the younger defer to the older through showing respect. \\
        \midrule
        {"topic": "SOCIAL VALUES", "q\_id": "28", "option": "1. Strongly agree 2. agree 3. Disagree 4. Strongly disagree", "q\_content": "When a mother works for pay, the children suffer"} & The culture of Bengal defines the cultural heritage of the Bengali people native to eastern regions of the Indian subcontinent, mainly what is today Bangladesh and the Indian states of West Bengal and Tripura, where they form the dominant ethnolinguistic group and the Bengali language is the official and primary language. Bengal has a recorded history of 1,400 years. After the partition, Bangladeshi culture became distinct from the mainstream Bengali culture. & (Spain - Value) Flexibility in social interactions and a relaxed approach to time and financial contributions among peers. \\
        \midrule
        {"topic": "SOCIAL VALUES", "q\_id": "29", "option": "1. Strongly agree 2. agree 3. Disagree 4. Strongly disagree", "q\_content": "On the whole, men make better political leaders than women do"} & Chinese culture is one of the world's oldest cultures, originating thousands of years ago. The culture prevails across a large geographical region in East Asia with Sinosphere in whole and is extremely diverse, with customs and traditions varying greatly between counties, provinces, cities, towns. The terms 'China' and the geographical landmass of 'China' have shifted across the centuries, before the name 'China' became commonplace in modernity. & (UK - Story) Elena was planning to visit her friend Tom's house. She texted Tom to ask if it was a good time for him. Upon arrival, she made sure not to overstay by asking if he needed her to leave at a certain time. Is what Elena did socially acceptable? \\
        \bottomrule
    \end{tabular}
    \caption{Comparison of data across different cultures from the data sources used in our paper.}
    \label{apptab:data_samples}
\end{table*}

\section{Training Procedure and Data Reformulation}
\label{app:training_reformulation}

Following \citet{cultureLLM/corr/abs-2402-10946}, our experiments employ LoRA adapters with 4-bit quantization using the BitsAndBytes configuration to optimize the memory usage. We use an alpha value of 16, a dropout rate of 0.1, and a rank of 64, specifically targeting the query (\texttt{q\_proj}) and value (\texttt{v\_proj}) projection matrices of the transformer architecture.

We reformulate the training data using the following formats:

\begin{enumerate}[noitemsep,topsep=0pt]
    \item \textbf{Standard Survey Training (WVS).} The WVS survey data is structured with clear task markers:

    \begin{verbatim}
    ### Task: Survey Question-Answer
    ### Question: [question_content]
    ### Answer: [answer_content]
    \end{verbatim}
    
    \item \textbf{Wikipedia.} When the Wikipedia data is used, the information is formatted as:
    
    \begin{verbatim}
    ### Task: Cultural Context
    ### Culture: [culture_name]
    ### Description: [cultural_context]
    \end{verbatim}

    \item \textbf{NormAd.} We integrate the data using the following prompt:
    
    \begin{verbatim}
    ### Task: NormAd Cultural Context
    ### Culture: [culture_name]
    ### Country: [country_name]
    ### Background: [background_info]
    ### Rule-of-Thumb: [cultural_rule]
    ### Story: [narrative]
    ### Explanation: [detailed_explanation]
    \end{verbatim}
\end{enumerate}

The training process optimizes memory usage with gradient checkpointing and uses a constant learning rate of \(2 \times 10^{-4}\). The model is trained for 6 epochs with a warmup ratio of 0.03 and employs 8-bit Adam optimization with a weight decay of 0.001. For reproducibility, the process is seeded (\texttt{seed=42}) and ensures deterministic CUDA operations.

\clearpage
\section{Full Performance Tables}
\label{app:full_perf_table}

\subsection{Zero-Shot Prompting and Single Culture Adaptation Results}

\begin{table*}[ht!]
\centering
\footnotesize
\begin{tabular}{lrrrrrrrrrrr}
\toprule
\textbf{Model} & \textbf{ara} & \textbf{ben} & \textbf{zho} & \textbf{eng} & \textbf{deu} & \textbf{ell} & \textbf{kor} & \textbf{por} & \textbf{spa} & \textbf{tur} & \textbf{Avg.} \\
\midrule
\multicolumn{11}{c}{\textbf{Zero-Shot Prompting}} \\
\midrule
Llama-3.1-8B & 11.96 & 17.12 & 32.77 & 14.85 & 23.81 & 38.16 & 26.14 & 19.93 & 30.96 & 21.95 & 23.77 \\
Llama-3.1-8B-IT & 19.14 & 23.10 & 30.49 & 26.63 & 34.36 & 37.56 & 38.72 & 20.92 & 39.14 & 32.95 & 30.00 \\
Gemma-2-9b-IT & 17.98 & 50.65 & 20.30 & 46.30 & 50.18 & 45.94 & 60.40 & 38.80 & 27.40 & 46.35 & 40.43 \\
Qwen2.5-7B-Instruct & 45.41 & 58.88 & 25.30 & 38.29 & 60.30 & 48.27 & 53.86 & 54.87 & 45.72 & 60.37 & 49.13 \\
\midrule
\multicolumn{11}{c}{\textbf{Single-Culture Adaptation - WVS}} \\
\midrule
Llama-3.1-8B & 17.22 & 22.01 & 38.28 & 19.92 & 29.30 & 36.08 & 32.65 & 20.15 & 27.93 & 28.57 & 27.21 \\
Llama-3.1-8B-IT & 19.50 & 23.51 & 32.69 & 22.35 & 34.78 & 36.98 & 37.61 & 17.75 & 25.85 & 28.78 & 27.98 \\
Gemma-2-9b-IT & 15.54 & 43.95 & 24.10 & 33.92 & 41.01 & 49.09 & 61.01 & 37.66 & 37.15 & 48.81 & 39.22 \\
Qwen2.5-7B-Instruct & 39.30 & 59.24 & 25.78 & 40.39 & 57.85 & 48.02 & 53.79 & 51.77 & 51.31 & 57.47 & 48.49 \\
\bottomrule
\end{tabular}
\caption{Culture adaptation results (F1 scores) under three training scenarios: zero-shot prompting and single-culture adaptation (training on Llama-3.1-8B models using WVS data). Evaluation uses a multilingual offensiveness dataset (\S\ref{subsec:dataset}), reported as averaged F1 scores.}
\label{tab:combined-baseline-zs}
\end{table*}

\subsection{Full Invalid Ratio}
\label{app:full_invalid_ratio}

\begin{table}[h]
\centering
\footnotesize
\begin{tabular}{llrr}
\toprule
\textbf{Methods} &  & \textbf{Inv. Cult. (\%)} & \textbf{Inv. MMLU (\%)} \\
\midrule

\multirow{6}{*}{Llama-3.1-8B} 
  & Zero‑Shot            & 20.12 & 2.3 \\
  & WVS   & 14.68 & 0 \\
  & \textbf{NormAd}      & 15.90 & \textbf{70.0} \\
  & WVS+Wiki             & 14.04 & 0 \\
  & WVS+NormAd           & 13.22 & 0 \\
  & WVS+Wiki+NormAd    & 12.85 & 0 \\
\midrule
\multirow{6}{*}{Llama-3.1-8B‑IT}
  & Zero‑Shot            & 21.20 & 0 \\
  & WVS   & 10.82 & 0 \\
  & \textbf{NormAd}      & 11.73 & \textbf{72.3} \\
  & WVS+Wiki             &  9.73 & 0 \\
  & WVS+NormAd           &  8.91 & 0 \\
  & WVS+Wiki+NormAd    &  8.35 & 0 \\
\midrule
\multirow{6}{*}{Gemma‑2‑9B‑IT}
  & Zero‑Shot            &  13.23 & 0 \\
  & WVS   &  0 & 0 \\
  & \textbf{NormAd}      &  9.7 & \textbf{82.7} \\
  & WVS+Wiki             &  6.32 & 0 \\
  & WVS+NormAd           &  5.89 & 0 \\
  & WVS+Wiki+NormAd    &  6.21 & 0 \\
\midrule
\multirow{6}{*}{Qwen2.5‑7B‑IT}
  & Zero‑Shot            &  9.4 &  0 \\
  & WVS   &  0 &  0 \\
  & \textbf{NormAd}      &  7.5 & \textbf{10.1} \\
  & WVS+Wiki             &  0 &  0 \\
  & WVS+NormAd           &  0 &  0 \\
  & WVS+Wiki+NormAd    &  0 &  0 \\
\bottomrule
\end{tabular}
\caption{Invalid response rates on cultural evaluation sets (\textit{Invalid Cult.}) and on MMLU (\textit{Invalid MMLU}).  
All MMLU invalid ratios are lower than the 20.12 \% cultural baseline of Llama‑3.1‑8B—\emph{except} for the purposely inflated \textbf{NormAd}-only rows, which remain dramatically worse.}
\label{tab:invalid-responses-complete}
\end{table}

\subsection{Combined Cultural Adaptation}
Instead of learning a separate adapter per culture, we combine training data from all target cultures and produce one multi-culture adapter. This can potentially help the model recognize cross-cultural patterns or exploit data from many cultures. However, it risks ``averaging out'' the distinctions, possibly causing \textit{cultural interference} (e.g., losing the unique viewpoint for each culture, akin to interference in multilinguality \citealt{cursexling/conneau-etal-2020-unsupervised, cursexling/wang-etal-2020-negative}). While combined-culture adaptation can improve some low-resource cultures (e.g., Korean, Bengali), it could reduce performance for others, indicating cultural interference.

\begin{table*}[h!]
    \centering
    \resizebox{\textwidth}{!}{ 
        \begin{tabular}{lccccccccccc}
            \toprule
            \multicolumn{12}{c}{\textbf{Combined-Culture Adaptation - WVS}} \\
            \midrule
            \textbf{Model} & \textbf{ara} & \textbf{ben} & \textbf{zho} & \textbf{eng} & \textbf{deu} & \textbf{ell} & \textbf{kor} & \textbf{por} & \textbf{spa} & \textbf{tur} & \textbf{Avg.} \\
            \midrule
            Llama-3.1-8B & 33.44 & 23.24 & 28.39 & 17.12 & 36.75 & 15.11 & 37.09 & 17.88 & 25.62 & 39.29 & 27.39 \\
            Llama-3.1-8B-IT & 28.00 & 30.34 & 42.77 & 23.90 & 46.08 & 31.42 & 43.32 & 22.88 & 33.52 & 43.50 & 34.57 \\
            \bottomrule
        \end{tabular}
    }
    \caption{Results for Combined-Culture Adaptation on WVS.}
    \label{tab:combined_culture_wvs}
\end{table*}

\vspace{-15pt}
\subsection{Freeform Generation}

\subsubsection{Performance Heatmaps - Llama-3.1-8B}
\label{app:performance_heatmaps}
\begin{figure*}[h!]
\centering
\begin{subfigure}[b]{0.32\textwidth}
    \centering
    \includegraphics[width=\textwidth]{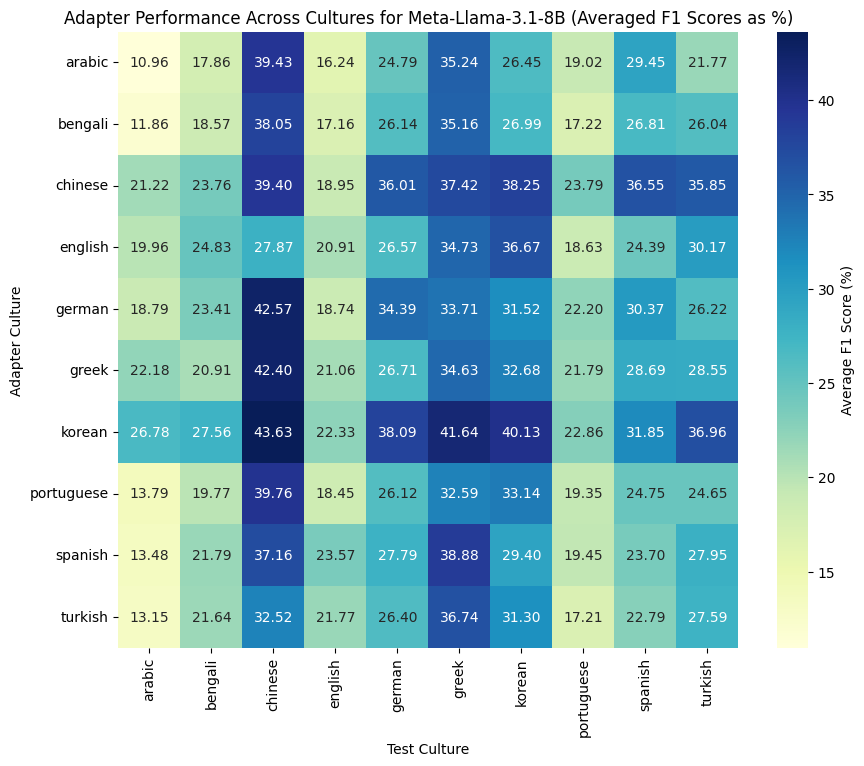}
    \caption{\textbf{WVS}}
\end{subfigure}
\hfill
\begin{subfigure}[b]{0.32\textwidth}
    \centering
    \includegraphics[width=\textwidth]{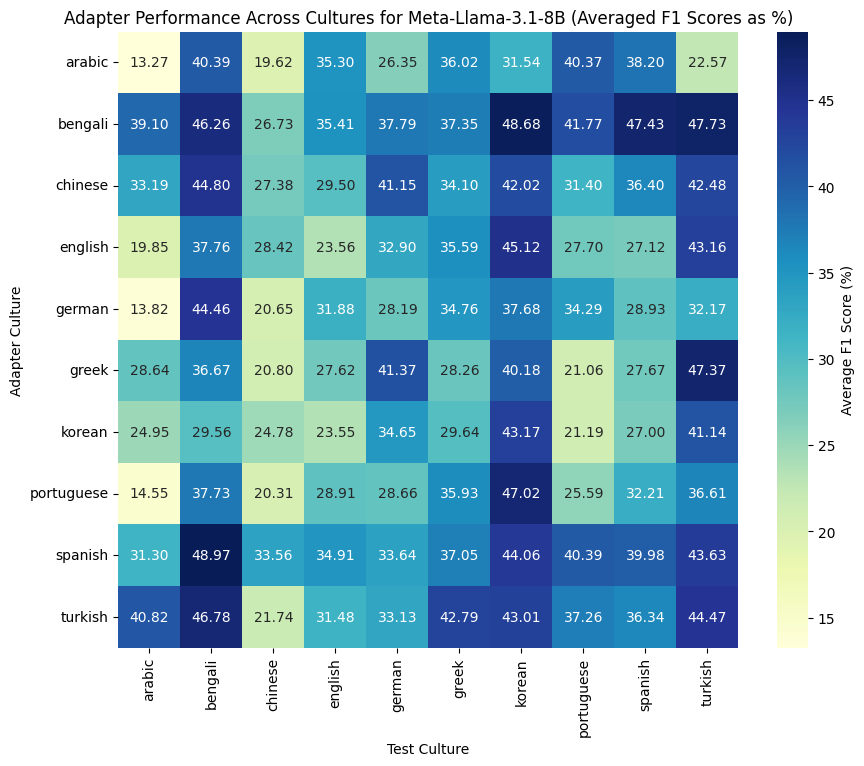}
    \caption{\textbf{WVS+Wiki}}
\end{subfigure}
\hfill
\begin{subfigure}[b]{0.32\textwidth}
    \centering
    \includegraphics[width=\textwidth]{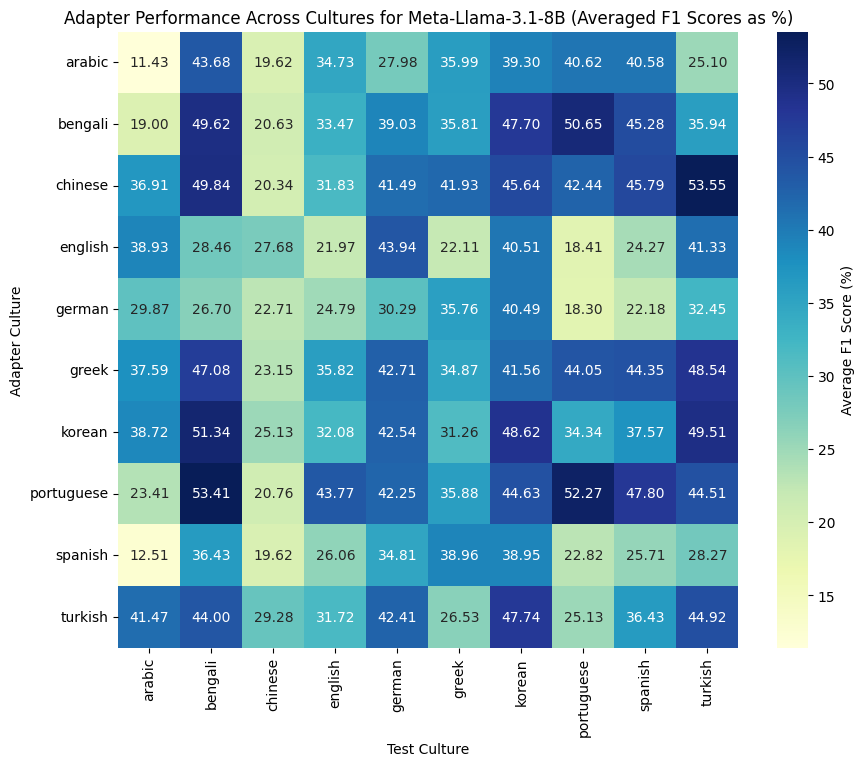}
    \caption{\textbf{WVS+NormAd}}
\end{subfigure}
\caption{Heatmaps of culture-specific classification performance (Llama-3.1-8B) using different data sources based on the ranks of the adaptation results.}
\label{appfig:three_clusters_llama}
\end{figure*}

\noindent Figure~\ref{appfig:three_clusters_llama} illustrates the culture-specific classification performance of the Llama-3.1-8B model through three heatmaps corresponding to different data configurations: panel (a) uses only WVS data, panel (b) integrates cultural context from Wikipedia (WVS+Wiki), and panel (c) combines WVS with NormAd data (WVS+NormAd); in each heatmap, color gradients represent the ranks of the adaptation results, providing a visual assessment of how incorporating additional cultural sources can enhance or alter model performance across diverse cultural contexts.

\clearpage
\subsubsection{Performance Tables - Llama-3.1-8B-Instruct}
\begin{figure*}[h!]
\centering
\begin{subfigure}[b]{0.32\textwidth}
    \centering
    \includegraphics[width=\textwidth]{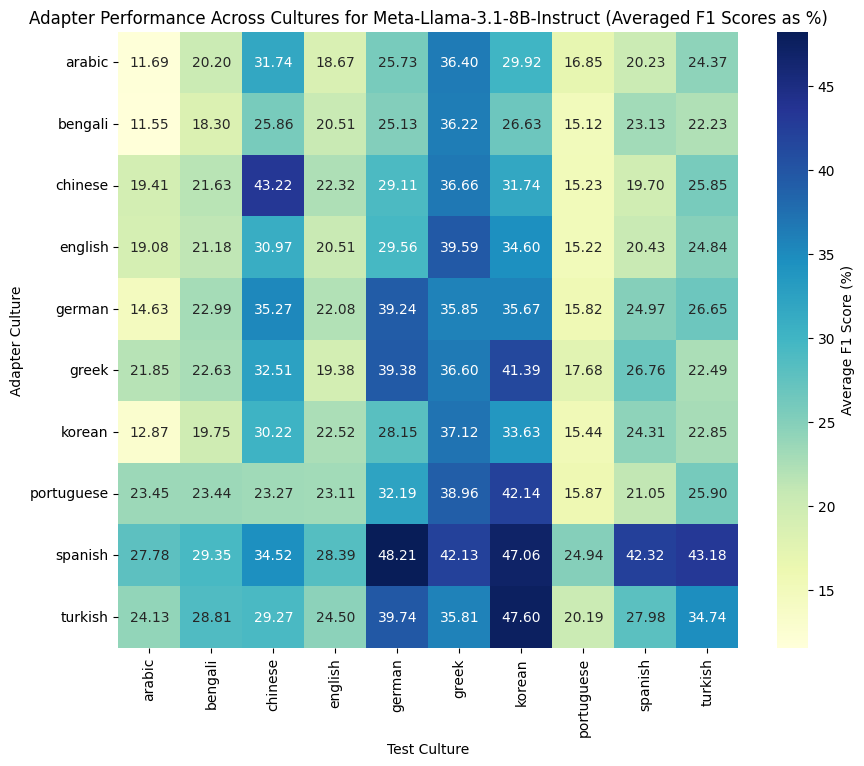}
    \caption{\textbf{WVS}}
\end{subfigure}
\hfill
\begin{subfigure}[b]{0.32\textwidth}
    \centering
    \includegraphics[width=\textwidth]{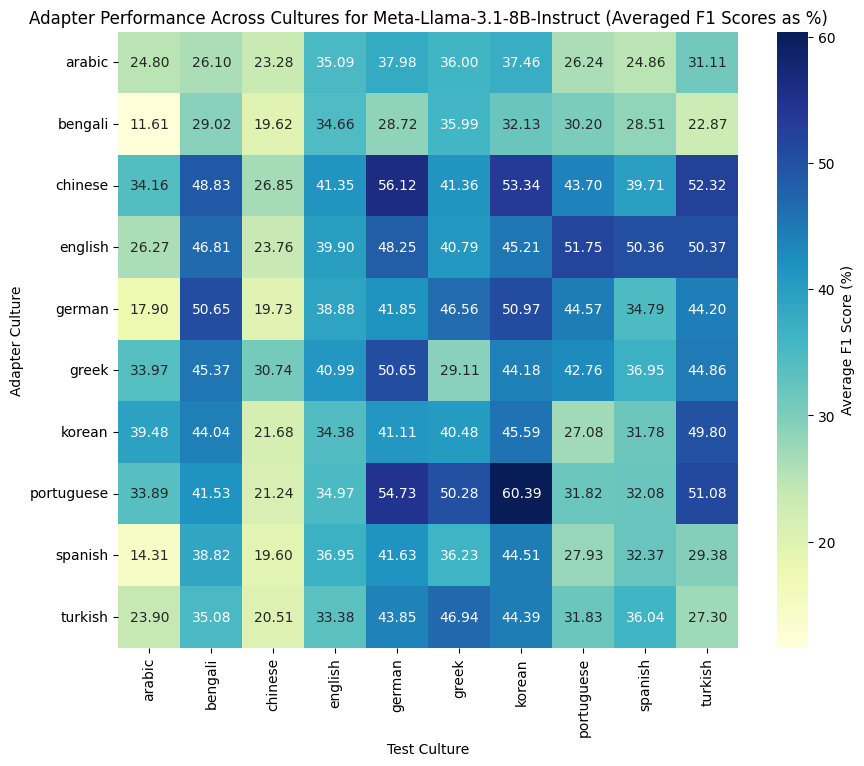}
    \caption{\textbf{WVS+Wiki}}
\end{subfigure}
\hfill
\begin{subfigure}[b]{0.32\textwidth}
    \centering
    \includegraphics[width=\textwidth]{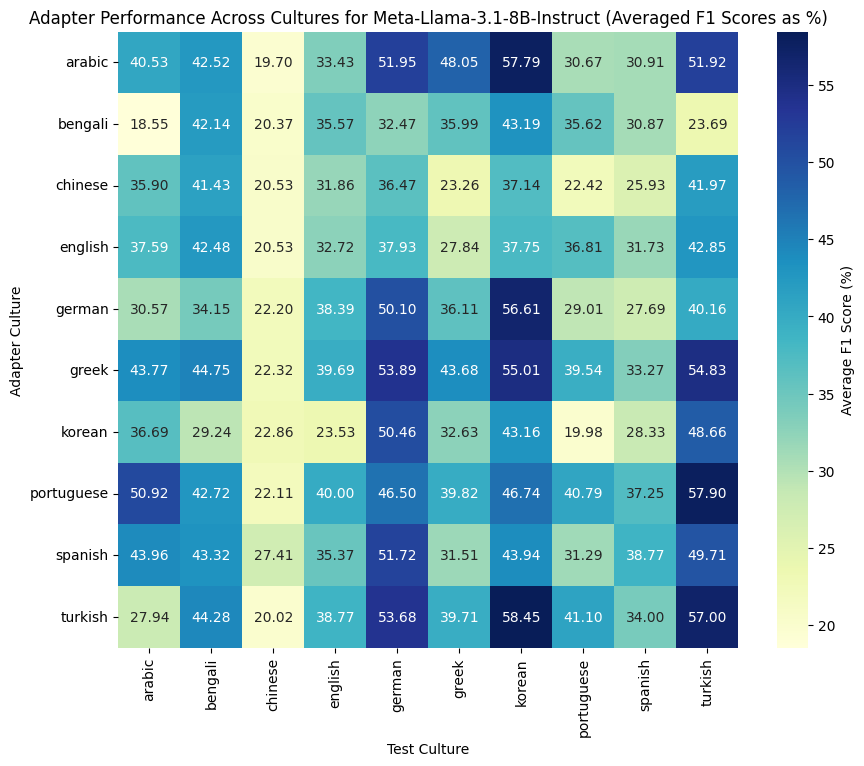}
    \caption{\textbf{WVS+NormAd}}
\end{subfigure}
\caption{Heatmaps of culture-specific classification performance (Llama-3.1-8B-IT) using different data sources based on the ranks of the adaptation results.}
\label{appfig:three_clusters_llamainst}
\end{figure*}

\noindent Figure~\ref{appfig:three_clusters_llamainst} illustrates the performance of Llama-3.1-8B-Instruct model through three heatmaps.

\vspace{-10pt}
\subsubsection{Performance Tables - Qwen2.5-7B-IT}
\begin{figure*}[h!]
\centering
\begin{subfigure}[b]{0.32\textwidth}
    \centering
    \includegraphics[width=\textwidth]{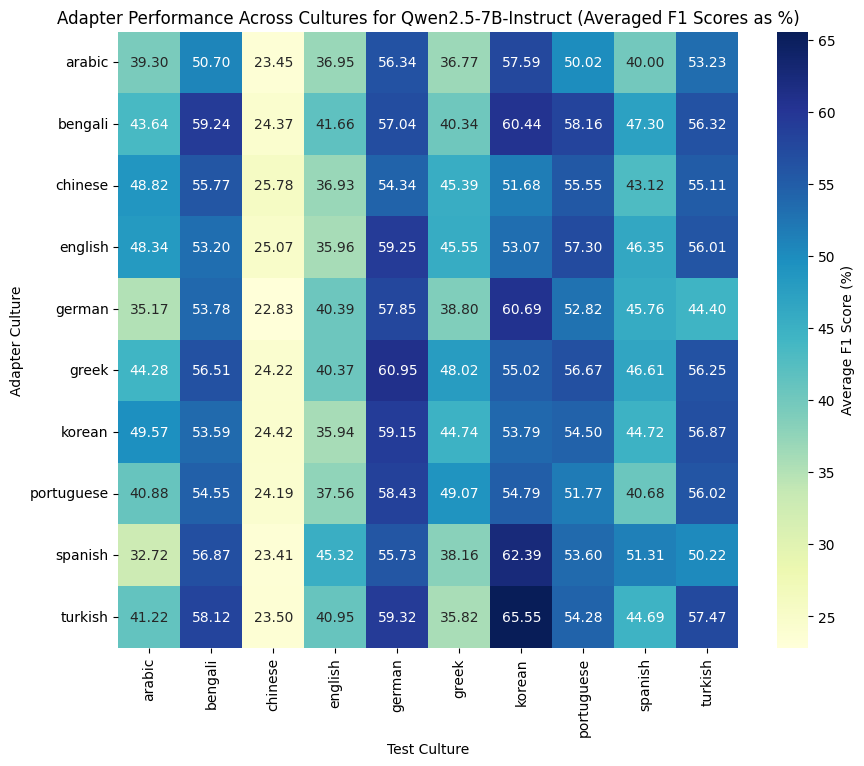}
    \caption{\textbf{WVS}}
\end{subfigure}
\hfill
\begin{subfigure}[b]{0.32\textwidth}
    \centering
    \includegraphics[width=\textwidth]{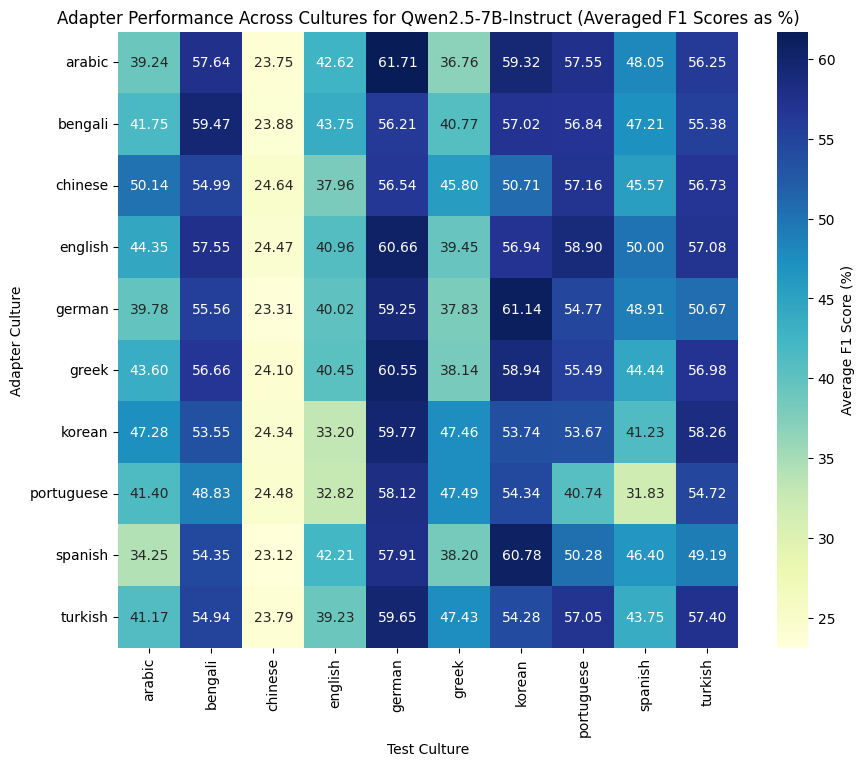}
    \caption{\textbf{WVS+Wiki}}
\end{subfigure}
\hfill
\begin{subfigure}[b]{0.32\textwidth}
    \centering
    \includegraphics[width=\textwidth]{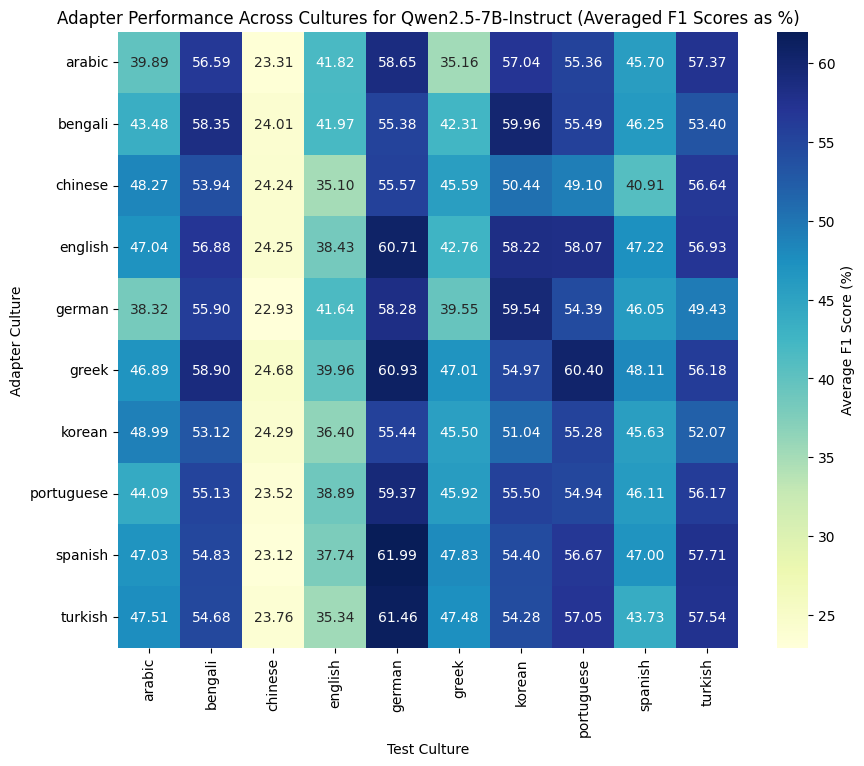}
    \caption{\textbf{WVS+NormAd}}
\end{subfigure}
\caption{Heatmaps of culture-specific classification performance (Qwen2.5-7B-IT) using different data sources based on the ranks of the adaptation results.}
\label{appfig:three_clusters_qwen}
\end{figure*}

\noindent Figure~\ref{appfig:three_clusters_qwen} illustrates the performance of the Qwen2.5-7B-IT model through three heatmaps.

\vspace{-10pt}
\subsubsection{Performance Tables - Gemma-2-9B-IT}
\begin{figure*}[h!]
\centering
\begin{subfigure}[b]{0.32\textwidth}
    \centering
    \includegraphics[width=\textwidth]{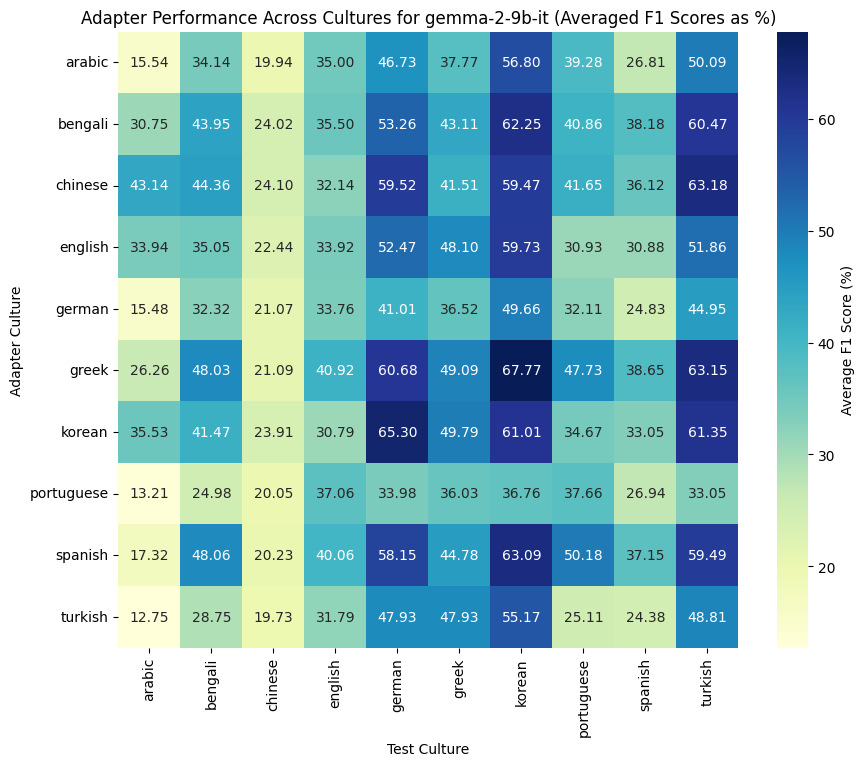}
    \caption{\textbf{WVS}}
\end{subfigure}
\hfill
\begin{subfigure}[b]{0.32\textwidth}
    \centering
    \includegraphics[width=\textwidth]{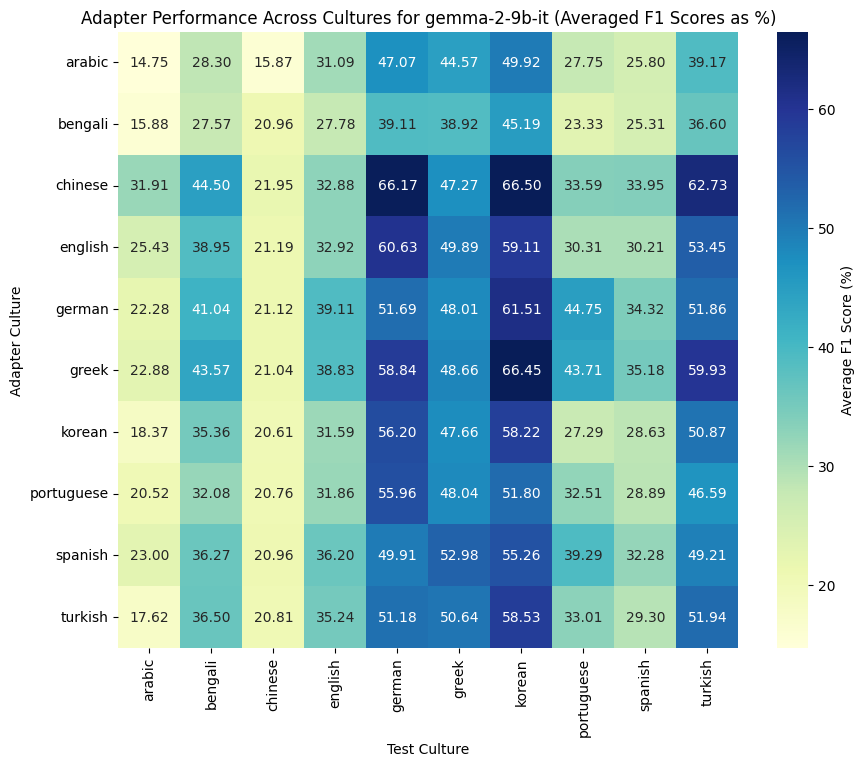}
    \caption{\textbf{WVS+Wiki}}
\end{subfigure}
\hfill
\begin{subfigure}[b]{0.32\textwidth}
    \centering
    \includegraphics[width=\textwidth]{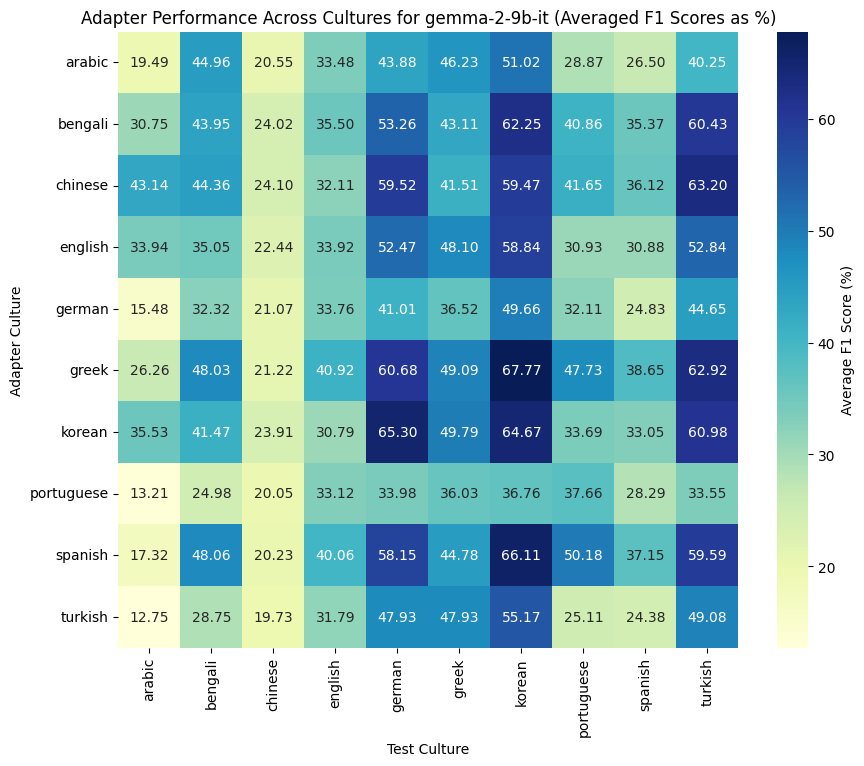}
    \caption{\textbf{WVS+NormAd}}
\end{subfigure}
\caption{Heatmaps of culture-specific classification performance (Gemma-2-9B-Instruct) using different data sources based on the ranks of the adaptation results.}
\label{appfig:three_clusters_gemma}
\end{figure*}

\noindent Figure~\ref{appfig:three_clusters_gemma} illustrates the performance of the Gemma-2-9B-IT model through three heatmaps.

\clearpage
\subsection{Normalized Scores Tables}
\label{app:normalized_scores}

\begin{table*}[ht!]
\centering
\footnotesize
\begin{tabular}{lrrrrrrrrrr}
\toprule
\textbf{Adapter Cult.} & \textbf{ara} & \textbf{ben} & \textbf{zho} & \textbf{eng} & \textbf{deu} & \textbf{ell} & \textbf{kor} & \textbf{por} & \textbf{spa} & \textbf{tur} \\
\midrule
\textbf{ara} & 0.4209 & 0.6882 & 0.7343 & 0.6578 & 0.5337 & 0.8640 & 0.6284 & 0.6758 & 0.4780 & 0.5645 \\
\textbf{ben} & 0.4156 & 0.6237 & 0.5984 & 0.7223 & 0.5213 & 0.8598 & 0.5595 & 0.6062 & 0.5466 & 0.5148 \\
\textbf{zho} & 0.6986 & 0.7371 & 1.0000 & 0.7862 & 0.6038 & 0.8703 & 0.6667 & 0.6107 & 0.4654 & 0.5985 \\
\textbf{eng} & 0.6867 & 0.7216 & 0.7166 & 0.7225 & 0.6131 & 0.9398 & 0.7268 & 0.6103 & 0.4828 & 0.5751 \\
\textbf{deu} & 0.5266 & 0.7835 & 0.8161 & 0.7779 & 0.8139 & 0.8509 & 0.7493 & 0.6345 & 0.5899 & 0.6172 \\
\textbf{ell} & 0.7865 & 0.7711 & 0.7522 & 0.6827 & 0.8168 & 0.8688 & 0.8695 & 0.7089 & 0.6324 & 0.5208 \\
\textbf{kor} & 0.4633 & 0.6728 & 0.6991 & 0.7933 & 0.5838 & 0.8810 & 0.7065 & 0.6193 & 0.5745 & 0.5292 \\
\textbf{por} & 0.8442 & 0.7987 & 0.5384 & 0.8142 & 0.6676 & 0.9248 & 0.8853 & 0.6364 & 0.4975 & 0.5997 \\
\textbf{spa} & 1.0000 & 1.0000 & 0.7987 & 1.0000 & 1.0000 & 1.0000 & 0.9886 & 1.0000 & 1.0000 & 1.0000 \\
\textbf{tur} & 0.8685 & 0.9817 & 0.6772 & 0.8628 & 0.8242 & 0.8501 & 1.0000 & 0.8094 & 0.6610 & 0.8045 \\
\bottomrule
\end{tabular}
\caption{Normalized Scores and \diagonality on Llama-3.1-8B-IT for WVS. Rows represent the adapter culture, and columns represent the culture test set.}
\label{tab:wvs}
\end{table*}

\begin{table*}[ht!]
\centering
\footnotesize
\begin{tabular}{lrrrrrrrrrr}
\toprule
\textbf{Adapter Cult.} & \textbf{ara} & \textbf{ben} & \textbf{zho} & \textbf{eng} & \textbf{deu} & \textbf{ell} & \textbf{kor} & \textbf{por} & \textbf{spa} & \textbf{tur} \\
\midrule
\textbf{ara} & 0.7255 & 0.5862 & 0.7980 & 0.8510 & 0.6329 & 0.7875 & 0.6219 & 0.7635 & 0.9012 & 0.5731 \\
\textbf{ben} & 0.3320 & 0.6027 & 0.4640 & 0.8319 & 0.5354 & 0.7861 & 0.5575 & 0.5934 & 0.7311 & 0.4903 \\
\textbf{zho} & 0.8268 & 0.7872 & 1.0000 & 0.9636 & 0.8755 & 1.0000 & 0.8753 & 0.8413 & 0.8521 & 0.7687 \\
\textbf{eng} & 0.7514 & 0.8592 & 0.9779 & 0.7852 & 0.9733 & 0.8209 & 0.9034 & 0.9299 & 0.9792 & 0.8828 \\
\textbf{deu} & 0.5986 & 0.8016 & 0.9445 & 0.7760 & 0.8604 & 0.9679 & 0.8233 & 0.7221 & 0.7729 & 0.6408 \\
\textbf{ell} & 0.9031 & 0.9440 & 0.7137 & 1.0000 & 0.9152 & 0.7502 & 0.8970 & 1.0000 & 1.0000 & 0.9678 \\
\textbf{kor} & 1.0000 & 1.0000 & 0.5369 & 0.8979 & 1.0000 & 0.8037 & 1.0000 & 0.8637 & 0.8274 & 1.0000 \\
\textbf{por} & 0.7863 & 0.7632 & 0.5586 & 0.8940 & 0.8065 & 0.9270 & 0.8570 & 0.7430 & 0.6613 & 0.7746 \\
\textbf{spa} & 0.4076 & 0.6871 & 0.5581 & 0.8136 & 0.6525 & 0.7973 & 0.7152 & 0.5486 & 0.6715 & 0.5138 \\
\textbf{tur} & 0.5835 & 0.6960 & 0.9223 & 0.8341 & 0.7417 & 0.8859 & 0.8456 & 0.7119 & 0.9690 & 0.6794 \\
\bottomrule
\end{tabular}
\caption{Normalized Scores and \diagonality on Llama-3.1-8B-IT for WVS+Wikipedia. Rows represent the adapter culture, and columns represent the culture test set.}
\label{tab:wvs_wikipedia}
\end{table*}

\begin{table*}[ht!]
\centering
\footnotesize
\begin{tabular}{lrrrrrrrrrr}
\toprule
\textbf{Adapter Cult.} & \textbf{ara} & \textbf{ben} & \textbf{zho} & \textbf{eng} & \textbf{deu} & \textbf{ell} & \textbf{kor} & \textbf{por} & \textbf{spa} & \textbf{tur} \\
\midrule
\textbf{ara} & 0.7961 & 0.8685 & 0.7190 & 0.8358 & 0.9640 & 1.0000 & 0.9533 & 0.7462 & 0.7974 & 0.8966 \\
\textbf{ben} & 0.3643 & 0.8608 & 0.7432 & 0.8893 & 0.6026 & 0.7490 & 0.7124 & 0.8666 & 0.7963 & 0.4092 \\
\textbf{zho} & 0.7051 & 0.8463 & 0.7493 & 0.7967 & 0.6767 & 0.4841 & 0.6127 & 0.5454 & 0.6689 & 0.7248 \\
\textbf{eng} & 0.7383 & 0.8678 & 0.7493 & 0.8180 & 0.7038 & 0.5794 & 0.6227 & 0.8956 & 0.8185 & 0.7400 \\
\textbf{deu} & 0.6004 & 0.6975 & 0.8100 & 0.9597 & 0.9297 & 0.7515 & 0.9337 & 0.7058 & 0.7142 & 0.6936 \\
\textbf{ell} & 0.8597 & 0.9141 & 0.8144 & 0.9923 & 1.0000 & 0.9091 & 0.9074 & 0.9620 & 0.8582 & 0.9469 \\
\textbf{kor} & 0.7207 & 0.5973 & 0.8340 & 0.5882 & 0.9363 & 0.6791 & 0.7118 & 0.4862 & 0.7307 & 0.8404 \\
\textbf{por} & 1.0000 & 0.8727 & 0.8067 & 1.0000 & 0.8628 & 0.8287 & 0.7709 & 0.9925 & 0.9607 & 1.0000 \\
\textbf{spa} & 0.8634 & 0.8849 & 1.0000 & 0.8843 & 0.9596 & 0.6558 & 0.7248 & 0.7613 & 1.0000 & 0.8585 \\
\textbf{tur} & 0.5487 & 0.9045 & 0.7305 & 0.9694 & 0.9960 & 0.8265 & 0.9640 & 1.0000 & 0.8771 & 0.9844 \\
\bottomrule
\end{tabular}
\caption{Normalized Scores and \diagonality on Llama-3.1-8B-IT for WVS+NormAd. Rows represent the adapter culture, and columns represent the culture test set.}
\label{tab:wvs_normad}
\end{table*}

\clearpage
\subsection{Probability-Based Generation}
\label{app:proba_based_eval}

Table \ref{tab:mmlu_wvs_proba} shows the normalized F1 score for probability-based generation evaluations.

\begin{table*}[h]
    \centering
    \small
    \begin{tabular}{l cc cc}
        \toprule
        \multirow{2}{*}{Language} & \multicolumn{2}{c}{Baseline} & \multicolumn{2}{c}{Translated} \\
                                 & Llama-3.1-8B & Llama-3.1-8B-IT & Llama-3.1-8B & Llama-3.1-8B-IT \\
        \midrule
        ara      & 30.52 & 28.83 & 33.24 & 37.81 \\
        ben     & 22.53 & 45.45 & 29.70 & 42.77 \\
        zho     & 28.84 & 41.35 & 35.77 & 46.28 \\
        eng     & 28.37 & 42.81 & 30.21 & 49.18 \\
        deu      & 32.53 & 40.40 & 28.80 & 41.92 \\
        ell       & 30.77 & 46.05 & 32.11 & 36.34 \\
        kor      & 30.28 & 41.80 & 34.33 & 44.63 \\
        por  & 29.24 & 40.11 & 27.55 & 38.08 \\
        spa     & 28.96 & 43.77 & 23.32 & 38.60 \\
        tur     & 30.44 & 43.93 & 30.24 & 40.46 \\
        \bottomrule
    \end{tabular}
    \caption{
        Performance on MMLU when training each adapter with different WVS cultural data.
        Baseline refers to fine-tuning using English-language cultural value data with the \textit{Llama-3.1-8B} and \textit{Llama-3.1-8B-IT} models.
        Translated represents training with WVS cultural values translated into the respective target language, using the \textit{Llama-3.1-8B} and \textit{Llama-3.1-8B-IT} models.
        The zero-shot performance for Arabic is 0.35 with \textit{Llama-3.1-8B} and 0.45 with \textit{Llama-3.1-8B-IT}.
    }
    \label{tab:mmlu_wvs_proba}
\end{table*}

\section{Invalid Answer Check}
\label{app:invalid_check}

\subsection{Code for Invalid Answer Filtering}
We process the generated response to determine whether it contains a valid answer using the following function. If the expected answer format is not detected, a default value is assigned.

\begin{lstlisting}[language=Python, caption=Generating Response and Invalid Response filtering from Model., label=code:get_response]
# Generate response from model
def getResponse(prompt, base_model=None, base_tokenizer=None):
    input_ids = base_tokenizer(prompt, return_tensors="pt").input_ids.to("cuda")

    base_model.eval()
    with torch.no_grad():
        output = base_model.generate(
            input_ids,
            max_new_tokens=25,
            temperature=0.0,
            num_beams=1,
            do_sample=False,
            eos_token_id=base_tokenizer.eos_token_id,
            pad_token_id=base_tokenizer.pad_token_id,
        )

    output_txt = base_tokenizer.decode(output[0], skip_special_tokens=True)

    # Extract the answer (e.g., "A")
    match = re.search(r'### Answer:\s*([A-D])', output_txt)
    if match:
        prediction = match.group(1)
        invalid_response = False
    else:
        prediction = "A"  # Default or handle as appropriate
        invalid_response = True

    return output_txt, prediction, invalid_response
\end{lstlisting}

\clearpage
\subsection{Invalid Examples}
\label{app:invalid_examples}
Table \ref{apptab:invalid-responses} illustrates examples of invalid responses from our evaluation datasets (English, German, and Spanish). The model failed to properly complete its task of detecting toxic content, showing truncated or incomplete outputs across different languages.

\begin{table}[h!]
\centering
\small
\begin{tabular}{p{1.5cm} | p{11.5cm}}
\toprule
\textbf{Culture} & \textbf{Invalid Response Example} \\
\midrule
English & \texttt{\{"input": "@user @user legit nilas retarded idk", "output": "@user @user legit nilas retarded idk what to do @user @user @user", "extracted\_output": "", "prediction": "", "label": "0", "invalid\_response": true\}} \\
\midrule
German & \texttt{\{"input": "@FobiMania @ManuelaSchwesig @MartinSchulz Ich scheisse auf die SPD. |LBR| SCHEINHEILGSTE PARTEI DEUTSCHLANDS !!!", "output": "\#\#\# Question: Is the following sentence offensive: @FobiMania @ManuelaSchwesig @MartinSchulz Ich scheisse auf die SPD. |LBR| SCHEINHEILGSTE PARTEI DEUTSCHLANDS!!! \textbackslash"1. Offensive\textbackslash", \textbackslash"2. Not offensive\textbackslash" without explanation. \#\#\# Answer: @FobiMania @ManuelaSchw", "extracted\_output": "@FobiMania @ManuelaSchw", "prediction": "@fobimania @manuelaschw", "label": "OFF", "invalid\_response": true\}} \\
\midrule
Spanish & \texttt{\{"input": "A ver si le dan un Goya al youtuber que se llevó la hostia por subnormal.", "output": "\#\#\# Question: Does the following sentence contain hate speech: A ver si le dan un Goya al youtuber que se llevó la hostia por subnormal. \textbackslash"1. Hatespeech\textbackslash", \textbackslash"2. Not Hatespeech\textbackslash" without explanation. \#\#\# Answer: A ver si le dan un Goya al y", "extracted\_output": "A ver si le dan un Goya al y", "prediction": "a ver si le dan un goya al y", "label": "HS", "invalid\_response": true\}} \\
\bottomrule
\end{tabular}
\caption{Examples of invalid responses.}
\label{apptab:invalid-responses}
\end{table}

\subsection{Invalid Rate}
\label{app:invalid_rate}

Table~\ref{tab:invalid_ratio} shows our findings in evaluating the zero-shot model, compared to the finetuned models on WVS (Single-Culture-WVS) and WVS+Wikipedia (Single-Culture-WVS+Wiki). This table shows that the zero-shot performance of the models still suffers from the lack of instruction capability, even on the instruct variant. We further see a notable decrease in the invalid response ratio after finetuning.

\begin{table*}[h]
    \footnotesize
    \centering
    \begin{tabular}{l l r r r}
        \toprule
        \textbf{Model} & \textbf{Scenario} & \textbf{Avg. Invalid Count} & \textbf{Total Samples} & \textbf{Invalid Ratio} \\
        \midrule
        \multirow{3}{*}{Llama-3.1-8B} & Zero-shot & 222.5849 & 1106 & 21.77\% \\
         & Single-Culture-WVS & 143.6604 & 1106 & 14.06\% \\
         & Single-Culture-WVS+Wiki & 133.6604 & 1106 & 12.97\% \\
         \multirow{3}{*}{Llama-3.1-8B-IT} & Zero-shot & 234.5660 & 1106 & 22.55\% \\
          & Single-Culture-WVS & 134.3774 & 1106 & 13.16\% \\
                                          & Single-Culture-WVS+Wiki & 121.5849 & 1106 & 12.01\% \\                    
        \midrule
       \multirow{3}{*}{Llama-3.1-8B}  & Zero-shot & 11,797 & 58,638 & 20.12\% \\
                                    & Single-Culture-WVS & 94,663 & 645,018 & 14.68\% \\
                                    & Single-Culture-WVS+Wiki & 90,562 & 645,018 & 14.04\% \\
        \multirow{3}{*}{Llama-3.1-8B-IT}    & Zero-shot & 12,432 & 58,638 & 21.20\% \\
                                            & Single-Culture-WVS & 69,804 & 645,018 & 10.82\% \\
                                          & Single-Culture-WVS+Wiki & 71,019 & 645,018 & 11.01\% \\
        \bottomrule
    \end{tabular}
    \caption{Invalid count and ratio for different models and scenarios.}
    \label{tab:invalid_ratio}
\end{table*}

\clearpage
\section{Prompts}
\label{app:prompts}

Table \ref{tab:prompt-format} summarizes the prompt templates used by the \texttt{getPrompt} function for various tasks.

\begin{table*}[h!]
\centering
\footnotesize
\begin{tabular}{p{3.7cm}p{11.4cm}}
\toprule
\textbf{Task(s)} & \textbf{Prompt Format} \\
\midrule

\texttt{offensive\_detect}, \texttt{abusive\_detect}
&
\texttt{\#\#\# Question: is the following sentence offensive: \{input\_txt\}.} \newline
\texttt{"1. Offensive", "2. Not offensive" without explanation.} \newline
\texttt{\#\#\# Answer:}
\\
\midrule

\texttt{hate\_detect} (excluding \texttt{hate\_detect\_fine-grained})
&
\texttt{\#\#\# Question: does the following sentence contain hate speech: \{input\_txt\}.} \newline
\texttt{"1. Hatespeech", "2. Not Hatespeech" without explanation.} \newline
\texttt{\#\#\# Answer:}
\\
\midrule

\texttt{vulgar\_detect\_mp}
&
\texttt{\#\#\# Question: does the following sentence contain vulgar speech: \{input\_txt\}.} \newline
\texttt{"1. Vulgar", "2. Not Vulgar" without explanation.} \newline
\texttt{\#\#\# Answer:}
\\
\midrule

\texttt{spam\_detect}
&
\texttt{\#\#\# Question: is the following sentence a spam tweet: \{input\_txt\}.} \newline
\texttt{"1. Spam", "2. Not Spam" without explanation.} \newline
\texttt{\#\#\# Answer:}
\\
\midrule

\texttt{hate\_detect\_fine-grained}
&
\texttt{\#\#\# Question:} \newline
\texttt{Does the following sentence contain hate speech?} \newline
\texttt{\{input\_txt\}} \newline
\texttt{Please choose one of the following options without explanation:} \newline
\texttt{1. Not Hatespeech}, \newline
\texttt{2. Race}, \newline
\texttt{3. Religion}, \newline
\texttt{4. Ideology}, \newline
\texttt{5. Disability}, \newline
\texttt{6. Social Class},
\texttt{7. Gender}, \newline
\texttt{\#\#\# Answer:}
\\
\midrule

\makecell[l]{\texttt{offensive\_detect}\\\texttt{finegrained}}
&
\texttt{\#\#\# Question:} \newline
\texttt{Does the following sentence contain offensive speech?} \newline
\texttt{\{input\_txt\}} \newline
\texttt{Please choose one of the following options without explanation:} \newline
\texttt{1. Not hatespeech} \newline
\texttt{2. Profanity, or non-targeted offense} \newline
\texttt{3. Offense towards a group} \newline
\texttt{4. Offense towards an individual} \newline
\texttt{5. Offense towards an other (non-human) entity} \newline
\texttt{\#\#\# Answer:}
\\
\midrule

\texttt{hate\_off\_detect}
&
\texttt{\#\#\# Question: does the following sentence contain hate speech or offensive content:} \newline
\texttt{\{input\_txt\}. "1. Hate or Offensive", "2. Not Hate or Offensive"} \newline
\texttt{without explanation.} \newline
\texttt{\#\#\# Answer:}
\\
\midrule

\makecell[l]{
\texttt{stereotype\_detect}, \\\texttt{mockery\_detect},\\
\texttt{insult\_detect}, \\\texttt{improper\_detect},\\
\texttt{aggressiveness\_detect}, \\\texttt{toxicity\_detect},\\
\texttt{negative\_stance\_detect}, \\\texttt{homophobia\_detect},\\
\texttt{racism\_detect}, \\\texttt{misogyny\_detect},\\
\texttt{threat\_detect}, \\\texttt{hostility\_directness\_detect}}
&
\texttt{\#\#\# Question: does the following sentence contain \{entity\}: \{input\_txt\}.} \newline
\texttt{"0. No", "1. Yes" without explanation.} \newline
\texttt{\#\#\# Answer:} \newline
(\textit{Note:} \texttt{\{entity\}} is derived from the task name, e.g., \texttt{bias\_on\_gender\_detect} $\rightarrow$ \textit{gender bias}, etc.)
\\
\midrule

\texttt{hate\_offens\_detect}
&
\texttt{\#\#\# Question: does the following sentence contain hate speech:} \newline
\texttt{\{input\_txt\}. "0. No", "1. Yes" without explanation.} \newline
\texttt{\#\#\# Answer:}
\\
\bottomrule
\end{tabular}

\caption{Overview of prompts generated by \texttt{getPrompt}.}
\label{tab:prompt-format}

\end{table*}

\clearpage
\section{Data Statistics}
\label{app:statistics}

\subsection{Training Data Statistics}
\label{app:train_data_source}

Table \ref{tab:sources_urls} lists the data sources and URLs utilized in our experiments, encompassing the World Values Survey (WVS), Wikipedia cultural articles, and the NormAd dataset. Tables \ref{tab:wiki_stats} and \ref{tab:normad_stats} provide detailed summary statistics for the Wikipedia and NormAd datasets respectively, outlining the total number of sentences, samples, and tokens per language.

\begin{table}[h!]
    \centering
    \small
    \begin{tabular}{@{}ll@{}}
    \toprule
    \textbf{Source} & \textbf{URL} \\
    \midrule
    World Values Survey (WVS) & \href{https://www.worldvaluessurvey.org/wvs.jsp}{WVS} \\
    Wikipedia (Arab Culture) & \href{https://en.wikipedia.org/wiki/Arab_culture}{Arab Culture} \\
    Wikipedia (Bengal Culture) & \href{https://en.wikipedia.org/wiki/Culture_of_Bengal}{Culture of Bengal} \\
    Wikipedia (Chinese Culture) & \href{https://en.wikipedia.org/wiki/Chinese_culture}{Chinese Culture} \\
    Wikipedia (English Culture) & \href{https://en.wikipedia.org/wiki/Culture_of_England}{Culture of England} \\
    Wikipedia (German Culture) & \href{https://en.wikipedia.org/wiki/Culture_of_Germany}{Culture of Germany} \\
    Wikipedia (Greek Culture) & \href{https://en.wikipedia.org/wiki/Culture_of_Greece}{Culture of Greece} \\
    Wikipedia (Korean Culture) & \href{https://en.wikipedia.org/wiki/Culture_of_Korea}{Culture of Korea} \\
    Wikipedia (Portuguese Culture) & \href{https://en.wikipedia.org/wiki/Culture_of_Portugal}{Culture of Portugal} \\
    Wikipedia (Spanish Culture) & \href{https://en.wikipedia.org/wiki/Culture_of_Spain}{Culture of Spain} \\
    Wikipedia (Turkish Culture) & \href{https://en.wikipedia.org/wiki/Culture_of_Turkey}{Culture of Turkey} \\
    NormAd Dataset & \href{https://huggingface.co/datasets/akhilayerukola/NormAd}{NormAd} \\
    \bottomrule
    \end{tabular} 
    \caption{Data sources and URLs.}
    \label{tab:sources_urls}
\end{table}

\begin{table}[h!]
    \centering
    \small
    \begin{tabular}{@{}lrrr@{}}
    \toprule
    \textbf{Language} & \textbf{Total Sentences} & \textbf{Total Tokens (Entire Text)} & \textbf{Total Tokens (Summed per Sentence)} \\
    \midrule
    Arabic & 257 & 8,990 & 9,018 \\
    Bengali & 127 & 4,282 & 4,307 \\
    Chinese & 388 & 13,929 & 13,938 \\
    English & 434 & 15,632 & 15,688 \\
    German & 171 & 6,322 & 6,338 \\
    Greek & 250 & 11,806 & 11,825 \\
    Korean & 150 & 5,678 & 5,687 \\
    Portuguese & 186 & 10,286 & 10,298 \\
    Spanish & 76 & 3,662 & 3,666 \\
    Turkish & 143 & 6,573 & 6,581 \\
    \bottomrule
    \end{tabular}
    \caption{Summary statistics for each language in our Wikipedia training dataset.}
    \label{tab:wiki_stats}
\end{table}

\begin{table}[h!]
    \centering
    \small
    \begin{tabular}{@{}lrr@{}}
    \toprule
    \textbf{Language} & \textbf{Samples} & \textbf{Tokens} \\
    \midrule
    Arabic & 239 & 102,705 \\
    Spanish & 234 & 74,674 \\
    Chinese & 134 & 35,988 \\
    English & 209 & 82,144 \\
    Korean & 27 & 6,784 \\
    German & 76 & 21,209 \\
    Bengali & 33 & 7,659 \\
    Portuguese & 77 & 19,022 \\
    Greek & 69 & 23,961 \\
    Turkish & 35 & 15,391 \\
    \bottomrule
    \end{tabular}
    \caption{Summary statistics for each language in our NormAd training dataset.}
    \label{tab:normad_stats}
\end{table}

\subsection{Test Data Statistics}
\label{app:test_data_statistics}
Following \citet{cultureLLM/corr/abs-2402-10946}, we break down our culture test set in the table below.

\begin{table*}[t!]
\label{tb-datasets}
\resizebox{\textwidth}{!}{
\begin{tabular}{c|c|c|c}
\toprule
Culture & Country \& Territory & Task \& Dataset & \#Sample \\ \midrule
\makecell{Arabic\\(\method-Ar)}  & Middle East & \makecell{\textit{Offensive language detection:} OffensEval2020(2000)~\citep{zampieri2020semeval}, \\OSACT4(1000)~\citep{husain2005osact4}, \\Multi-Platform(1000)~\citep{chowdhury2020multi}, \\and OSACT5(2541)~\citep{mubarak2022overview}. \\\textit{Hate detection:} OSACT4(1000)~\citep{husain2005osact4}, \\Multi-Platform(675)~\citep{chowdhury2020multi}, \\OSACT5(2541)~\citep{mubarak2022overview}, \\and OSACT5\_finegrained(2541)~\citep{mubarak2022overview}. \\\textit{Spam detection:} ASHT(1000)~\citep{kaddoura2024dataset}. \\\textit{Vulgar detection:} Multi-Platform(675)~\citep{chowdhury2020multi}}    & 14,973 \\ \midrule
\makecell{Bangli\\(\method-Bn)}  & Bangladesh & \makecell{\textit{Offensive language detection:} TRAC2020 Task1(1000)~\citep{trac2-dataset}, \\TRAC2020 Task2(1000)~\citep{trac2-dataset}, \\BAD(1000)~\citep{sharif2022tackling}. \\\textit{Hate detection:} Hate Speech(1000)~\citep{romim2021hate}. \\\textit{Threat detection:} BACD(1000)~\citep{Bangla-Abusive-Comment-Dataset}. \\\textit{Bias detection:} BACD(1000)~\citep{Bangla-Abusive-Comment-Dataset}.}     & 6,000 \\ \midrule
\makecell{Chinese\\(\method-Zh)}  & China & \makecell{\textit{Spam detection:} CCS(1000)~\citep{jiang2019detect}. \\\textit{Bias detection:} CDial-Bias(1000)~\citep{zhou2022towards}. \\\textit{Stance detection:} CValues(1712)~\citep{xu2023cvalues}.}  & 3,712 \\ \midrule
\makecell{English\\(\method-En)}  & United States & \makecell{\textit{Offensive language detection:} SOLID(1000)~\citep{rosenthal2020solid}. \\\textit{Hate detection:} MLMA(1000)~\citep{ousidhoum-etal-multilingual-hate-speech-2019} \\and HOF(1000)~\citep{hateoffensive}. \\\textit{Threat detection:} CValuesJMT(1000)~\citep{Jigsaw-Multilingual-Toxicity}.\\\textit{Toxicity detection:} MLMA(1000)~\citep{ousidhoum-etal-multilingual-hate-speech-2019} \\and JMT(1000)~\citep{Jigsaw-Multilingual-Toxicity}.}  & 6,000 \\ \midrule
\makecell{German\\(\method-De)}  & \makecell{Germany and \\parts of Europe} & \makecell{\textit{Offensive language detection:} GermEval2018(3531)~\citep{wiegand2018overview}. \\\textit{Hate detection:} IWG\_1(469)~\citep{ross2016hatespeech}, \\IWG\_2(469)~\citep{ross2016hatespeech}, HASOC2020(850)~\citep{HASOC2020}, \\and multilingual-hatecheck(1000)~\citep{rottger2022multilingual}.}  & 6,319 \\ \midrule
\makecell{Korean\\(\method-Ko)}  & South Korea & \makecell{\textit{Hate detection:} K-MHaS(1000)~\citep{lee-etal-2022-k}, \\hateSpeech(1000)~\citep{moon-etal-2020-beep}, \\and HateSpeech2(1000)~\citep{Korean-HateSpeech-dataset}. \\\textit{Abusive detection:} AbuseEval(1000)~\citep{caselli2020feel}, \\CADD(1000)~\citep{song2021large}, \\and Waseem(1000)~\citep{waseem2016hateful}.}  & 5,000 \\ \midrule
\makecell{Portuguese\\(\method-Pt)} & \makecell{Brazil and \\parts of \\Latin America} & \makecell{\textit{Offensive language detection:} OffComBR(1250)~\citep{Pelle2017}, \\and HateBR(1000)~\citep{vargas-etal-2022-hatebr}. \\\textit{Bias detection:} ToLD-Br-homophobia(1000)~\citep{leite2020toxic}, \\and ToLD-Br-misogyny(1000)~\citep{leite2020toxic}. \\\textit{Abusive detection:} ToLD-Br-insult(1000)~\citep{leite2020toxic}.}  & 16,250 \\ \midrule
\makecell{Spanish\\(\method-Es)} & \makecell{Argentina, \\Mexico, \\and parts of \\Latin America} & \makecell{\textit{Offensive language detection:} AMI(1000)~\citep{fersini2018overview}, \\MEX-A3T(1000)~\citep{alvarez2018overview}, \\and OffendES(1000)~\citep{plaza2021offendes}. \\\textit{Hate detection:} HatEval 2019(1000)~\citep{basile2019semeval}, \\and HaterNet(1000)~\citep{pereira2019detecting}. \\\textit{Bias detection:} DETOXIS\_stereotype(1000)~\citep{de2021ai}, \\and DETOXIS\_improper(1000)~\citep{de2021ai}. \\\textit{Abusive detection:} DETOXIS\_abusive(1000)~\citep{de2021ai}, \\DETOXIS\_mockery(1000)~\citep{de2021ai}. \\\textit{Aggressiveness detection:} DETOXIS\_aggressiveness(1000)~\citep{de2021ai}. \\\textit{Stance detection:} DETOXIS\_stance(1000)~\citep{de2021ai}.}  & 11,000 \\ \midrule
\makecell{Turkish\\(\method-Tr)} & Turkey & \makecell{\textit{Offensive language detection:} SemEval-2020(3528)~\citep{zampieri2020semeval}, \\offenseCorpus(1000)~\citep{coltekin2020lrec}, \\offenseKaggle(1000)~\citep{turkish-tweets}, \\and offenseKaggle\_2(1000)~\citep{turkish-offensive-language-detection}. \\\textit{Abusive detection:} ATC(1000)~\citep{karayiugit2021detecting}. \\\textit{Spam detection:} Turkish Spam(825)~\citep{misc_turkish_spam_v01_530}. \\\textit{Fine-grained offensive detection:} offenseCorpus(1000)~\citep{coltekin2020lrec}.}  & 10,353 \\ \midrule
                                         
\end{tabular}
}
\caption{Overview of the eight evaluation tasks and the 59 datasets used, including dataset names and their corresponding test sample sizes. For example, "OffensEval2020(2000)~\citep{zampieri2020semeval}" indicates that the OffensEval2020 dataset contains 2,000 test samples. }

\end{table*}

\clearpage

\section{Cross-Cultural Confusion Matrix on Llama-3.1-8B}
\begin{figure}[h]
    \centering
    \includegraphics[width=0.7\linewidth]{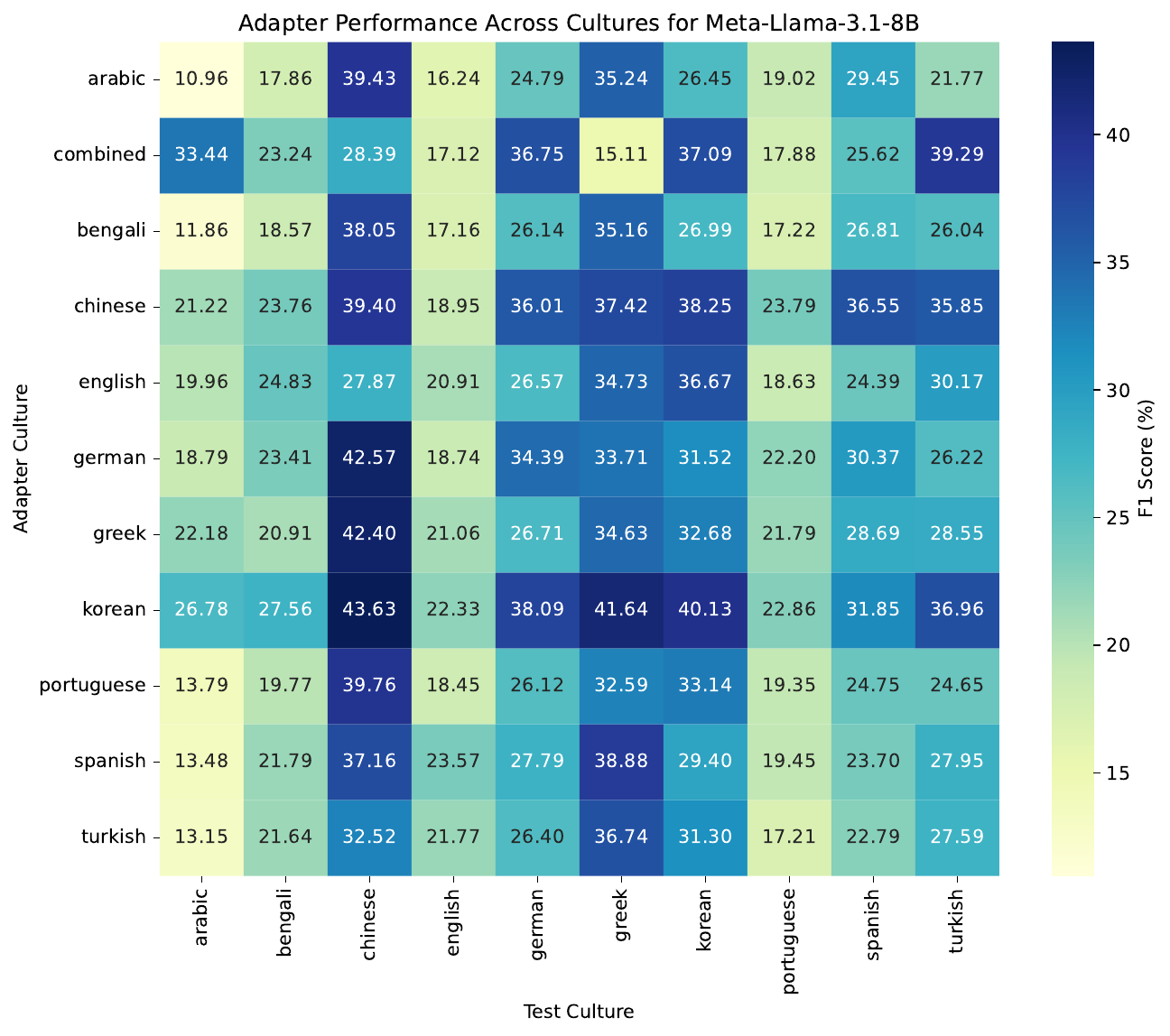}
    \caption{Cross-culture confusion matrix for the WVS-only baseline on Llama-3.1-8B (8B, base). The \diagonality score is $\approx 0.78$, reflecting substantial overlap in predictions across cultures.}
    \label{fig:no_diagonal_app}
\end{figure}

\end{document}